\documentclass[10pt,twocolumn,letterpaper]{article}

\usepackage{cvpr}              

\usepackage{graphicx}
\usepackage{multirow}
\usepackage{tikz}

\usepackage{amsmath,amsfonts,bm}

\def\eqref#1{equation~\ref{#1}}

\def\1{\bm{1}}

\DeclareMathAlphabet{\mathsfit}{\encodingdefault}{\sfdefault}{m}{sl}
\SetMathAlphabet{\mathsfit}{bold}{\encodingdefault}{\sfdefault}{bx}{n}

\usepackage{xcolor}
\usepackage[utf8]{inputenc} 
\usepackage[T1]{fontenc}    
\usepackage{url}            
\usepackage{booktabs}       
\usepackage{amsfonts}       
\usepackage{nicefrac}       
\usepackage{microtype}      
\usepackage{xcolor}       
\usepackage{xkcdcolors}
\usepackage{times}
\usepackage{epsfig}
\usepackage{graphicx}
\usepackage{placeins}
\usepackage{multirow}
\usepackage[skip=5pt]{caption}
\usepackage{rotating}
\usepackage{cancel}
\usepackage{setspace}

\usepackage{bm}
\usepackage{bibunits} 

\usepackage{amssymb,amsthm}
\usepackage{amsmath}
\usepackage{graphicx}
\usepackage{wrapfig,lipsum,booktabs}

\usepackage{epsfig}
\usepackage{tikz}
\usetikzlibrary{spy}
\usepackage{algpseudocode}
\usepackage{algorithm}
\usepackage{mathrsfs}

\usepackage{nicefrac}       
\usepackage{booktabs}       

\usepackage{thmtools,thm-restate}

\usepackage{listings}
\usepackage{lstautogobble}  %
\usepackage{color}          %
\usepackage{zi4}            %
\definecolor{bluekeywords}{rgb}{0.13, 0.13, 1}
\definecolor{greencomments}{rgb}{0, 0.5, 0}
\definecolor{redstrings}{rgb}{0.9, 0, 0}
\definecolor{graynumbers}{rgb}{0.5, 0.5, 0.5}
\definecolor{LightCyan}{rgb}{0.8, 0.9, 1}

\usepackage{subcaption}        
\usepackage{siunitx}               
\usepackage[export]{adjustbox}
\usepackage{makecell}
\usepackage{url}
\usepackage{tikz, pgfplots}
\usetikzlibrary{positioning}
\usepackage{capt-of}
\usepackage{diagbox}

\declaretheorem[]{definition}

\def\eqref#1{(\ref{#1})}

\usepackage{colortbl}
\usepackage{mdframed}

\makeatletter
\renewcommand{\paragraph}{%
  \@startsection{paragraph}{4}%
  {\z@}{1ex \@plus 1ex \@minus .2ex}{-1em}%
  {\normalfont\normalsize\bfseries}%
}
\makeatother

\definecolor{cvprblue}{rgb}{0.21,0.49,0.74}
\usepackage[pagebackref,breaklinks,colorlinks,allcolors=cvprblue]{hyperref}

\usepackage[accsupp]{axessibility}

\def\paperID{12175} 
\def\confName{CVPR}
\def\confYear{2025}

\title{Classifier-Free Guidance inside the Attraction Basin May Cause Memorization}

\author{Anubhav Jain$^{1}$\thanks{Work done during an internship at Sony AI.}, Yuya Kobayashi$^{2}$, Takashi Shibuya$^{2}$, Yuhta Takida$^{2}$,\\ Nasir Memon$^{1}$, Julian Togelius$^{1}$, Yuki Mitsufuji$^{2,3}$ \\
$^1$New York University, $^2$Sony AI, $^3$Sony Group Corporation\\
\texttt{\{aj3281,memon,julian.togelius\}@nyu.edu}\\ \texttt{\{u.kobayashi,takashi.tak.shibuya,yuta.takida,yuhki.mitsufuji\}@sony.com}
}

\newcommand{\ldenoiser}{denoiser}
\newcommand{\cdenoiser}{Denoiser}

\begin{document}

\maketitle

\begin{abstract}

Diffusion models are prone to exactly reproduce images from the training data. This exact reproduction of the training data is concerning as it can lead to copyright infringement and/or leakage of privacy-sensitive information. 
In this paper, we present a novel perspective on the memorization phenomenon and propose a simple yet effective approach to mitigate it.
We argue that memorization occurs because of an attraction basin in the denoising process which steers the diffusion trajectory towards a memorized image. However, this can be mitigated by guiding the diffusion trajectory away from the attraction basin by not applying classifier-free guidance until an ideal transition point occurs from which classifier-free guidance is applied. This leads to the generation of non-memorized images that are high in image quality and well-aligned with the conditioning mechanism. To further improve on this, we present a new guidance technique, \emph{opposite guidance}, that escapes the attraction basin sooner in the denoising process. We demonstrate the existence of attraction basins in various scenarios in which memorization occurs, and we show that our proposed approach successfully mitigates memorization. Our codebase is publicly available at \href{https://github.com/SonyResearch/mitigating_memorization}{https://github.com/SonyResearch/mitigating\_memorization}. 

\end{abstract}

\section{Introduction}

Recent advancements in image-generation models have enabled the generation of hyper-realistic images. Diffusion models \citep{song2020denoising,rombach2022high_LDM} have been at the forefront of this advancement with their ability to align image generation with different conditioning mechanisms such as text giving users more control over the output. 
However, a key challenge has emerged: these models can memorize and reproduce training examples verbatim (verbatim memorization) \citep{carlini2023extracting,somepalli2023diffusion_cvpr,somepalli2023understanding_neurips}, leading to concerns over copyright infringement and leakage of privacy sensitive information. Alternatively, the model can generate images copying the composition and structure present in one or more training images (template memorization)\cite{ren2024unveiling}. Researchers have identified factors contributing to memorization and reproduction, such as the duplication of training images and repeated captions \citep{somepalli2023understanding_neurips}. However, not all causes are understood, as memorization can still occur even when these issues are avoided. Therefore, there is a growing need for effective mitigation strategies, in particular methods that can be applied during inference without requiring computationally intensive retraining of the core diffusion models. 

\begin{figure}
    \centering
    \includegraphics[width=1.0\linewidth]{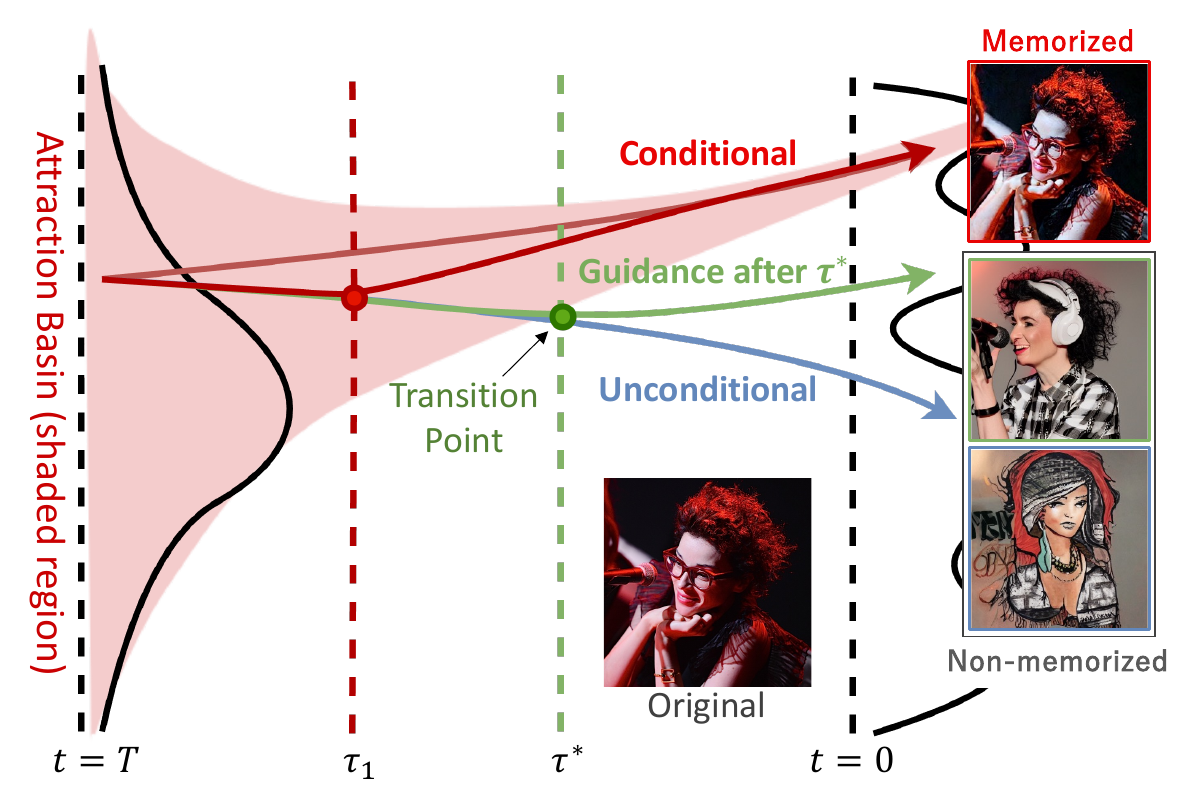}
    \caption{The diffusion trajectory contains an attraction basin (red region) which steers conditioned samples towards their memorized images. It can be avoided by applying zero classifier-free guidance when the trajectory is inside the attraction basin, such that there is an ideal transition point $\tau^*$ after which applying CFG leads to non-memorized output. Applying CFG at an earlier point such as $\tau_1$ inside the attraction basin leads to the same memorized sample. }
    \label{fig:attraction_basin_main}
\end{figure}

\begin{figure*}[htp]
    \centering
    \begin{subfigure}[t]{0.49\textwidth}
        \centering
        \includegraphics[width=\textwidth]{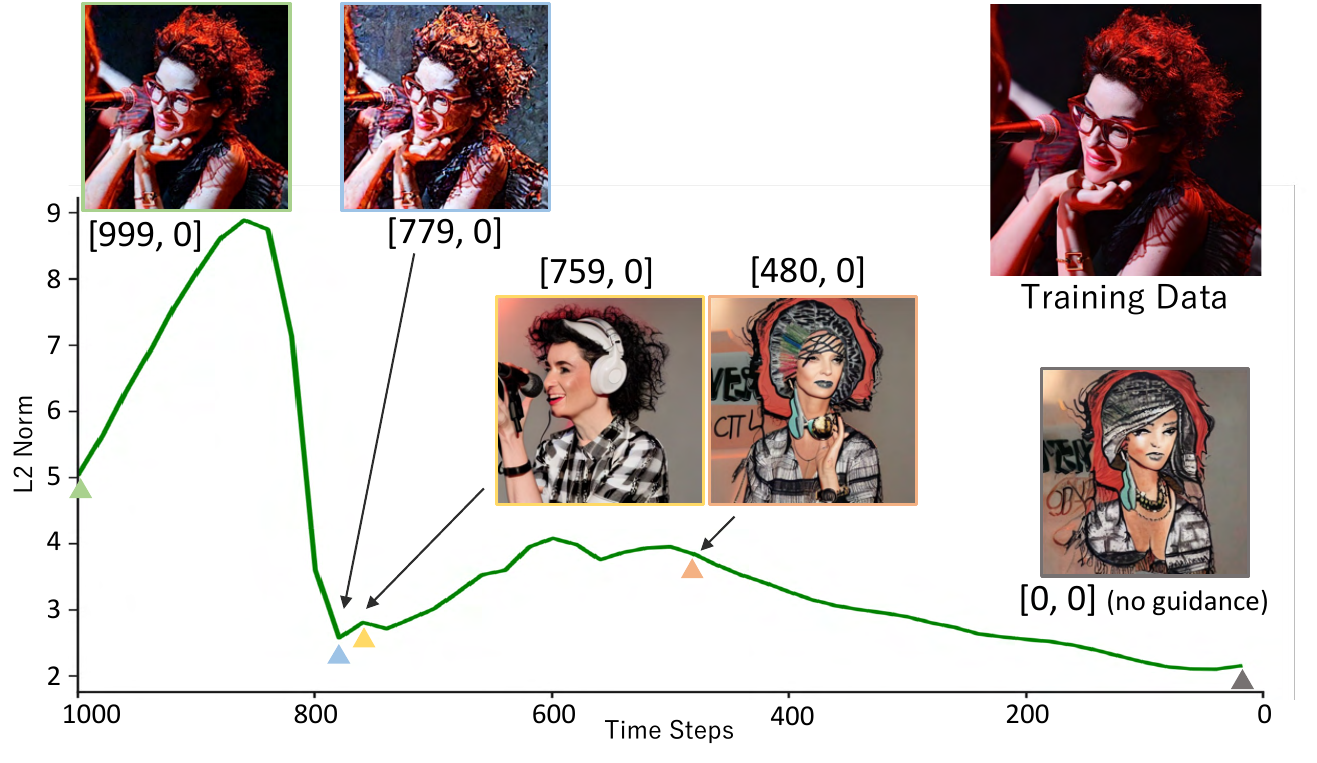}
        \caption{Prompt: ``Here's What You Need to Know About St. Vincent's Apple Music Radio Show''}
    \end{subfigure}
    ~
    \begin{subfigure}[t]{0.49\textwidth}
        \centering
        \includegraphics[width=\textwidth]{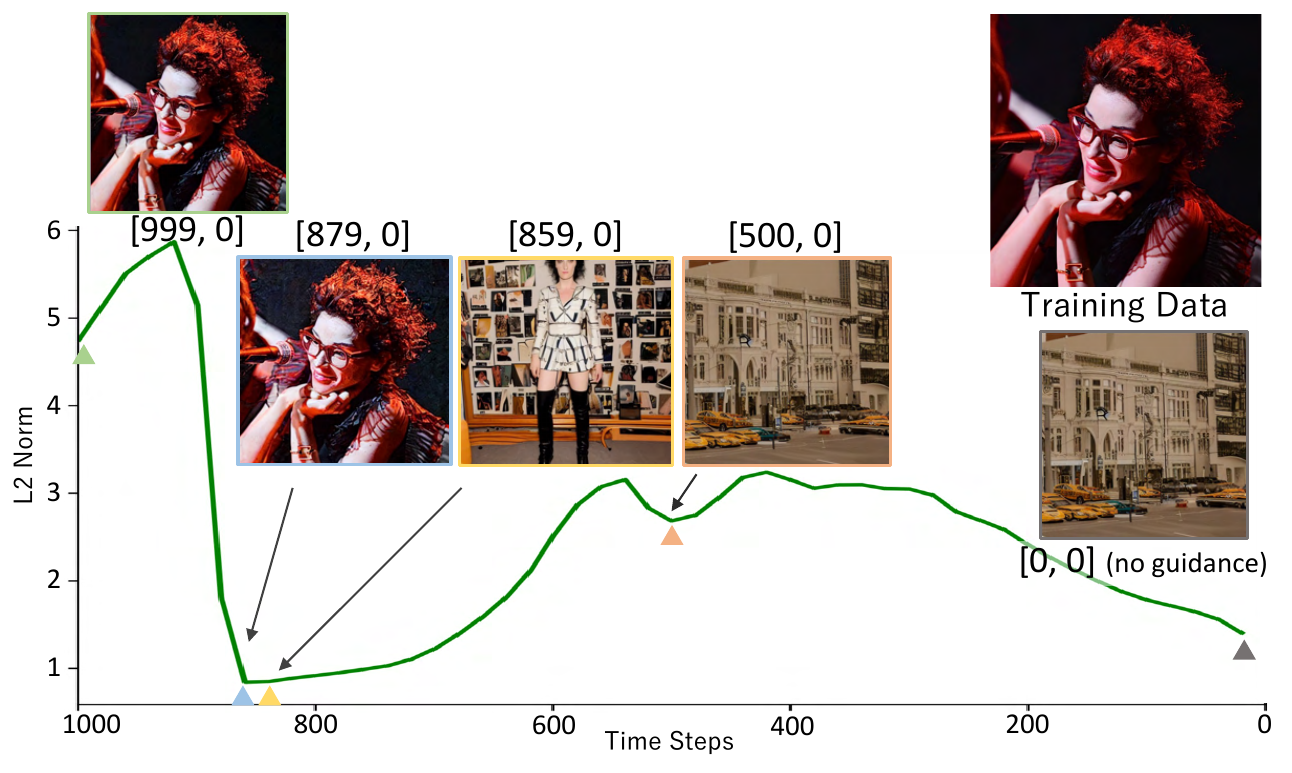}
        \caption{Prompt: ``Here's What You Need to Know About St. Vincent's Apple Music Radio Show''}
    \end{subfigure}
    \\
    \caption{Plots showing magnitude of $\epsilon_{\theta}(x_t, e_p) - \epsilon_{\theta}(x_t, e_{\emptyset})$ when denoising without classifier-free guidance (CFG) at each time step. The figures show the generated image if you start applying CFG at that time step. We get non-memorized output if we apply CFG after the ideal transition point $\tau^*$ which coincides with the fall in the conditional noise prediction. This value is dependent on both the prompt and the initialization ((a) and (b) contain the same prompt with different initializations). More examples in the Appendix.}
    \label{fig:attraction_basin}
    
\end{figure*}

In text-to-image (T2I) diffusion models, strong associations between a text prompt and an image in the training dataset lead to memorization. This can occur due to the presence of "trigger tokens" in overly specific text-prompts \citep{ren2024unveiling,wen2024detecting}, the model seeing multiple occurrences of these during training (data duplication) \citep{somepalli2023understanding_neurips,wen2024detecting}, or other causes such as fine-tuning on smaller datasets \citep{somepalli2023understanding_neurips}. Previous approaches have focused on perturbing either the text prompt/embeddings \citep{wen2024detecting,somepalli2023understanding_neurips} or the cross-attention between the text and image embeddings \citep{ren2024unveiling}. While these approaches help weaken the associations between prompts and corresponding images, they are limited to specific conditions in which they have been tested and fail to generalize effectively. Additionally, since these techniques work at the prompt level, they are only applicable to text-conditional generation, and also assume the ability to update the prompt.

In this paper, we introduce a new concept to explain the behavior of memorization in diffusion models based on our empirical observations. We first observe that applying classifier-free guidance (CFG) before a certain time step tends to produce memorized samples; we refer to this time step as the transition point. Interestingly, applying CFG after the transition point is unlikely to yield a memorized image.
Based on these observations, we propose the concept of an attraction basin, in which applying CFG induces a strong steering force toward a memorized output, even with different initializations. Furthermore, following the approach of Wen et al.~\citep{wen2024detecting}, we analyze the magnitude of the conditional noise predictions and discover uncharacteristically large values within the attraction basin, providing empirical evidence of a strong steering force in this region. Once the transition point is passed, the strong steering force is no longer pervasive. This magnitude-based analysis motivates us to detect the transition point by tracking the magnitude. Specifically, we initially apply no guidance until a transition point occurs —marked by a sharp drop in the discrepancy between conditional and unconditional guidance— after which applying CFG leads to non-memorized images.


We summarize our contributions in this paper as follows:
\begin{itemize}
    \item We present a new way to understand memorization in conditional diffusion models by understanding the diffusion trajectory from a dynamical systems theory perspective. We show that an attraction basin forms in the denoising trajectory. 
    \item We present approaches to avoiding the attraction basin in various scenarios that incur no additional computational costs and require neither access to the prompt nor modification of weights.
    \item We present a new guidance technique that we call Opposite Guidance (not to be confused with negative prompts~\cite{ban2024understanding}) to push the trajectory away from the attraction basin. 
    \item We comprehensively show that previous approaches are not generalizable in mitigating memorization and only work in certain scenarios studied in those papers. On the contrary, our approach is able to mitigate memorization in all the explored scenarios. 
\end{itemize}

\section{Related Work}

\subsection{Understanding Memorization}

In deep generative models, when memorization occurs, a generated sample can match verbatim with a single or set of training images (verbatim memorization) \citep{carlini2023extracting,somepalli2023diffusion_cvpr,somepalli2023understanding_neurips} or copy the same template as the training image (template memorization) \citep{ren2024unveiling} leading to privacy and/or copyright infringement issues. Researchers \cite{somepalli2023diffusion_cvpr, somepalli2023understanding_neurips,webster2023reproducible,carlini2023extracting} have identified that when images or prompts are duplicated in the training dataset, it can lead to memorization. Additionally, training on smaller datasets has also been shown to be a major factor in causing memorization \citep{somepalli2023understanding_neurips}. However, as Somepalli et al. \cite{somepalli2023understanding_neurips} pointed out, even though data has been de-duplicated in the newer models \citep{alexopenai} such as Stable Diffusion v2.1 (SDv2.1), the issue of memorization is yet to be resolved, and it does not explain much of the observed replication behavior. Factors such as conditioning and dataset complexity play a role even when the training datasets are large enough and are de-duplicated where "simpler" images and overly specific prompts have a higher chance of getting memorized \citep{somepalli2023understanding_neurips}. Thus, there is a strong need for mitigation techniques beyond dataset manipulations.   

\subsection{Detecting Memorization}

Similarity scores based on the SSCD (self-supervised copy detection) model \citep{pizzi2022self} can be used for getting an estimate on the extent of memorization wherein scores $>0.5$ can suggest that it is memorized \citep{somepalli2023diffusion_cvpr}. However, this metric is prone to false positives when the textual similarity is high between a generated and real image \citep{somepalli2023understanding_neurips}. Wen et al. \cite{wen2024detecting} showed that for memorized prompts, the difference between the text-conditioned and the unconditioned score predictions is uncharacteristically high and can be informative in detecting memorization. They demonstrated high performance even in the first step of noise prediction. Ren et al. \cite{ren2024unveiling}, on the other hand, showed that for memorized prompts, the cross-attention scores pertaining to certain "trigger tokens" in the memorized prompts are unusually high. Suggesting that they are given more importance, and thus lead to memorization. Other authors \cite{carlini2023extracting,webster2023reproducible} used a black box membership inference attack for finding memorized images. Daras et al. \cite{daras2024consistent} showed that when an image is memorized, the model is able to fully reconstruct it from a noised version of it. This observation could be used for detection. Another line of research has tried to understand memorization by comparing diffusion models to associative memory networks \cite{ambrogioni2024search,hoover2023memory,pham2024memorization}.

\subsection{Mitigating Memorization}

\subsubsection{Training Time Mitigation}

Wen et al. \cite{wen2024detecting} proposed monitoring the text-conditioned noise prediction scores for each sample and excluding it from the current mini-batch if it surpasses a certain \emph{pre-determined threshold}. On similar lines, Ren et al. \cite{ren2024unveiling} proposed removing samples from the mini-batch when their cross-attention entropy is above a particular \emph{pre-determined threshold}. Somepalli et al. \cite{somepalli2023understanding_neurips} proposed generating multiple BLIP captions per image during training. They also proposed perturbing the prompts by adding Gaussian noise or adding/replacing random words/numbers. Daras et al. \cite{daras2024consistent} proposed training/finetuning on corrupted images can reduce instances of data replication by the model. Chavhan et al. \cite{chavhan2024memorized} proposed pruning the UNet model weights assuming prior knowledge of memorized prompts. 

Most training time mitigations require re-training the base diffusion model from scratch. Not only is this computationally expensive, it also requires control over the entire training process including access to the training datasets.

\subsubsection{Inference Time Mitigation}

Wen et al. \cite{wen2024detecting} proposed perturbing the prompt embedding by minimizing the difference between the text-conditioned and unconditioned noise prediction scores such that it falls below a certain target loss. Ren et al. \cite{ren2024unveiling} proposed applying a mask and rescaling the attention scores to give less attention to "trigger tokens". However, this approach does not mitigate memorization when "trigger tokens" are not present. Somepalli et al. \cite{somepalli2023understanding_neurips} proposed adding random tokens to the text prompt or repeating certain tokens during inference so as to add noise to the text embedding indirectly. Chen et al. \cite{chen2024towards} proposed different CFG weight schedulers to mitigate different causes of memorization. The approach presumes knowledge of not only whether memorization is present but also the type of memorization. 

Most of these approaches work by updating text prompt/embeddings leads to a trade-off between the text alignment and memorization. Another drawback of these approaches is that they work better when a trigger token is present such that we can reduce the importance/weightage of that token and they do not generalize well to other types of memorization as we show in this paper. Lastly, since they work directly on the text prompt/embeddings, they cannot be applied to other conditioning mechanisms such as class labels.

\section{Preliminaries}

Diffusion models \citep{song2020denoising} such as Stable Diffusion (SD) \citep{rombach2022high_LDM} and Imagen \citep{saharia2022photorealistic} are trained with the objective of learning a model $\epsilon_{\theta}$ to denoise a noisy input vector at different levels of noise characterized by a time-dependent noise scheduler. We can define the diffusion process as a stochastic differential equation (SDE) in the form, 
\[
dx_t = f(x_t,t)dt + g(t)dw_t. 
\]
where \( x_t \in X \) is a sample at time \( t \), \( f(x_t,t) \) is the drift term, \( g(t) \) is the diffusion coefficient, and \( w_t \) is a standard Wiener process.

During training, the forward diffusion process comprises a Markov chain with fixed time steps $T$. Given a data point $x_0\sim q(x)$, we iteratively add Gaussian noise with variance $\beta_t$ at each time step $t$ to $x_{t-1}$ to get $x_t$ such that $x_T \sim \mathcal{N}(0, I)$. This process can be expressed as,
\[
    q(x_t | x_{t-1}) = \mathcal{N}(x_t; \sqrt{1-\beta_t}x_{t-1}, \beta_tI),\quad \forall t \in \{1, ..., T\}. \]

We can get a closed-form expression of $x_t$,
\begin{equation}
    \label{equation:forward_diffusion}
    x_t = \sqrt{\Bar{\alpha}_t}x_0 + \sqrt{1 - \Bar{\alpha}_t}\epsilon_t,
\end{equation}
where $\Bar{\alpha}_t = \prod_{i=1}^{t}(1-\beta_{i})$ and ${\alpha}_t=1-\beta_t$. 

During training, we learn the reverse process through a network $\epsilon_{\theta}$ to iteratively denoise $x_t$ by estimating the noise $\epsilon_t$ at each time step $t$. The loss function is expressed as, 
\begin{equation}
    \mathcal{L} = \mathbb{E}_{t \in [1,T],\epsilon \sim \mathcal{N}(0, I)} [\left\| \epsilon_t - \epsilon_{\theta} (x_t) \right\|_2^2].
\end{equation}

Using the learned noise estimator network $\epsilon_{\theta}$, we can compute the previous sample $x_{t-1}$ from $x_{t}$ as follows, 
\begin{equation}
    x_{t-1} =\frac{1}{\sqrt{\alpha}_t}(x_t - \frac{1- \alpha_t}{\sqrt{1-\Bar{\alpha_t}}}\epsilon_{\theta}(x_t)).
    \label{equation:x_t-1}
\end{equation}

The learned noise estimator network $\epsilon_{\theta}$ can be conditioned using a conditioning input such as text or class label, expressed as an embedding $e_p$. Ho et al. \cite{ho2022classifier} proposed classifier-free guidance (CFG) as a mechanism to guide the diffusion trajectory towards generating outputs that align with the conditioning. The trajectory is directed towards the conditional score predictions and away from the unconditional score predictions, where $s$ controls the degree of adjustment and $e_{\emptyset}$ are empty prompt embeddings used for unconditional guidance. 
\begin{equation}
    \hat{\epsilon} \leftarrow \epsilon_{\theta}(x_t, e_{\emptyset}) + s (\underbrace{\epsilon_{\theta}(x_t, e_p) - \epsilon_{\theta}(x_t, e_{\emptyset})}_{\text{conditional guidance}}).
    \label{eq:CFG}
\end{equation}

\section{Understanding Diffusion Trajectory \\ during Memorization}
\label{sec:understanding}

In text-to-image diffusion models, different initializations generally lead to different outputs for the same text prompt. However, when the diffusion model has memorized a particular sample, the outputs are similar and closely resemble one or more training data samples regardless of initialized noise \citep{wen2024detecting}.

As shown in Figure~\ref{fig:attraction_basin}, we observe that applying CFG after a certain time step avoids producing memorized outputs, while applying it from earlier time steps, including at the beginning ($t = T$), easily leads to memorized outputs.
Inspired by these empirical findings and the perspective of dynamical systems \cite{Milnor2006Attractor,auslander1964attractors}, we suspect that there is an attraction basin in the sample space of the diffusion trajectory, in which applying CFG results in memorized outputs.
Intuitively, one can imagine this attraction basin as having a funnel-like shape. It is broader at $t = T$ and narrows as the reverse denoising process progresses. To define the attraction basin, we introduce the concept of a \ldenoiser.
\begin{definition}[\cdenoiser]
    Let $X$ denote the entire sample space of the diffusion trajectory, and let $E$ denote the embedding space. A \ldenoiser{ }is a function \( \varphi: X\times(0,T]\times E \to X \) that outputs the result of the reverse diffusion process based on \cref{equation:x_t-1} with CFG. Specifically, \( \varphi(x,t,e) \) represents the resulting sample at time \( t=0 \), starting from a state \((x,t)\) with embeddings $e\in E$, where \( x \in X \) and $t\in(0,T]$.
\end{definition}
For an attractor $x^\mathrm{a}$ from the training dataset, we then define the attraction basin as follows: 
\begin{definition}[Attractor and attraction basin]\label{def:attraction_basin}
\color{black}

Suppose a prompt is given with embeddings denoted as $e_p$. A sample \( x^\mathrm{a} \in X \) is considered an ($\epsilon$-)attractor if the diffusion trajectories denoised with CFG tend to converge to samples within $B_{\mathcal{D}}(x^\mathrm{a},\epsilon)$ at $t=0$, regardless of the initial noise, i.e., $\varphi(x,T,e_p)\in B_{\mathcal{D}}(x^\mathrm{a},\epsilon)$ for any $x\in X$, where $B_{\mathcal{D}}(\cdot,\cdot)$ represents a perceptual ball given by a certain perceptual distance $\mathcal{D}:X\times X\to\mathbb{R}_{\geq0}$ as
\begin{align*}
    B_{\mathcal{D}}(x^\mathrm{a},\epsilon)=\left\{x'\in X\mid \mathcal{D}(x^\mathrm{a},x')\leq\epsilon\right\}.
\end{align*}
An attraction basin is a region in the state space within which samples will converge to around $x^\mathrm{a}$ at $t=0$ under CFG inference. More formally, the ($(\epsilon,\delta)$-)attraction basin defined using $x^\mathrm{a}$ is expressed as
\begin{align*}
X^\mathrm{b}(x^\mathrm{a}, \epsilon) &= \Bigl\{ (x,t) \mid \mathbb{P}(\varphi(x,t,e)\in B_{\mathcal{D}}(x^\mathrm{a},\epsilon)) > 1-\delta \Bigr\},
\end{align*}
representing all points in the state space from which generated samples using CFG will reach samples perceptually similar to the attractor at $t=0$. 
\color{black}
\end{definition}

Inspired from Wen et al.~\citep{wen2024detecting}, we analyze the attraction basin by examining the magnitudes of the conditional guidance, $\epsilon_{\theta}(x_t, e_p) - \epsilon_{\theta}(x_t, e_{\emptyset})$.
Interestingly, as shown in Figure \ref{fig:attraction_basin}, the conditional noise prediction $\epsilon_{\theta}(x_t, e_p) - \epsilon_{\theta}(x_t, e_{\emptyset})$ remains high during the initial time steps and then steeply falls at the time step immediately before the denoising trajectory exits the attraction basin. We refer to this time step as the transition point, marking the change from producing memorized outputs to non-memorized outputs.
In Figure \ref{fig:attraction_basin}(a), if you were to start applying CFG at any point on or before $t=779$, you get the same memorized output. However, surprisingly, applying CFG at exactly one denoising step after this $t=759$ leads to a non-memorized output (step size is 20 when denoising with 50 inference steps). We define the transition point as follows.  
\begin{definition}[Transition point]\label{def:transition_point}
\color{black}
Given prompt embeddings, let \( x^\mathrm{a} \in X \) be an attractor. We define a transition point as \( (x_{\tau}, \tau) \) for \( x_{\tau} \in X \) and $\tau>0$ such that \( (x_{\tau}, \tau+1) \) lies within its attraction basin, while \( (x_{\tau}, \tau) \) does not. More formally, an ($(\epsilon,\delta)$-)transition point satisfies
\begin{align*}
     &\mathbb{P}(\varphi(x_{\tau},\tau+1) \in B_{\mathcal{D}}(x^\mathrm{a},\epsilon)) > 1 - \delta \quad\text{and}\\
     &\mathbb{P}(\varphi(x_{\tau},\tau) \not\in B_{\mathcal{D}}(x^\mathrm{a},\epsilon)) < 1-\delta,
\end{align*}
where $B_\mathcal{D}$ is a perceptual ball defined in Definition~\ref{def:attraction_basin}. 
\color{black}
\end{definition}
This phenomenon occurs only for memorized samples as illustrated in Figure \ref{fig:dtpm_og_plots}, the conditional guidance $\epsilon_{\theta}(x_t, e_p) - \epsilon_{\theta}(x_t, e_{\emptyset})$ remains low and almost flat across different time steps for non-memorized samples. We further analyze the diffusion trajectory during memorization, demonstrating the presence of an attraction basin in Appendix \ref{sec:attraction_basin_extra_appendix}.

\section{Simple Mitigation Strategy}

Based on the intuition from the previous section, we can simply mitigate memorization by finding an ideal transition point to switch from zero CFG to applying CFG. Based on our preliminary experiments, we found that transition points can either be static or dynamic.

\begin{figure}[tbh]
\centering
    \includegraphics[width=0.7\columnwidth]{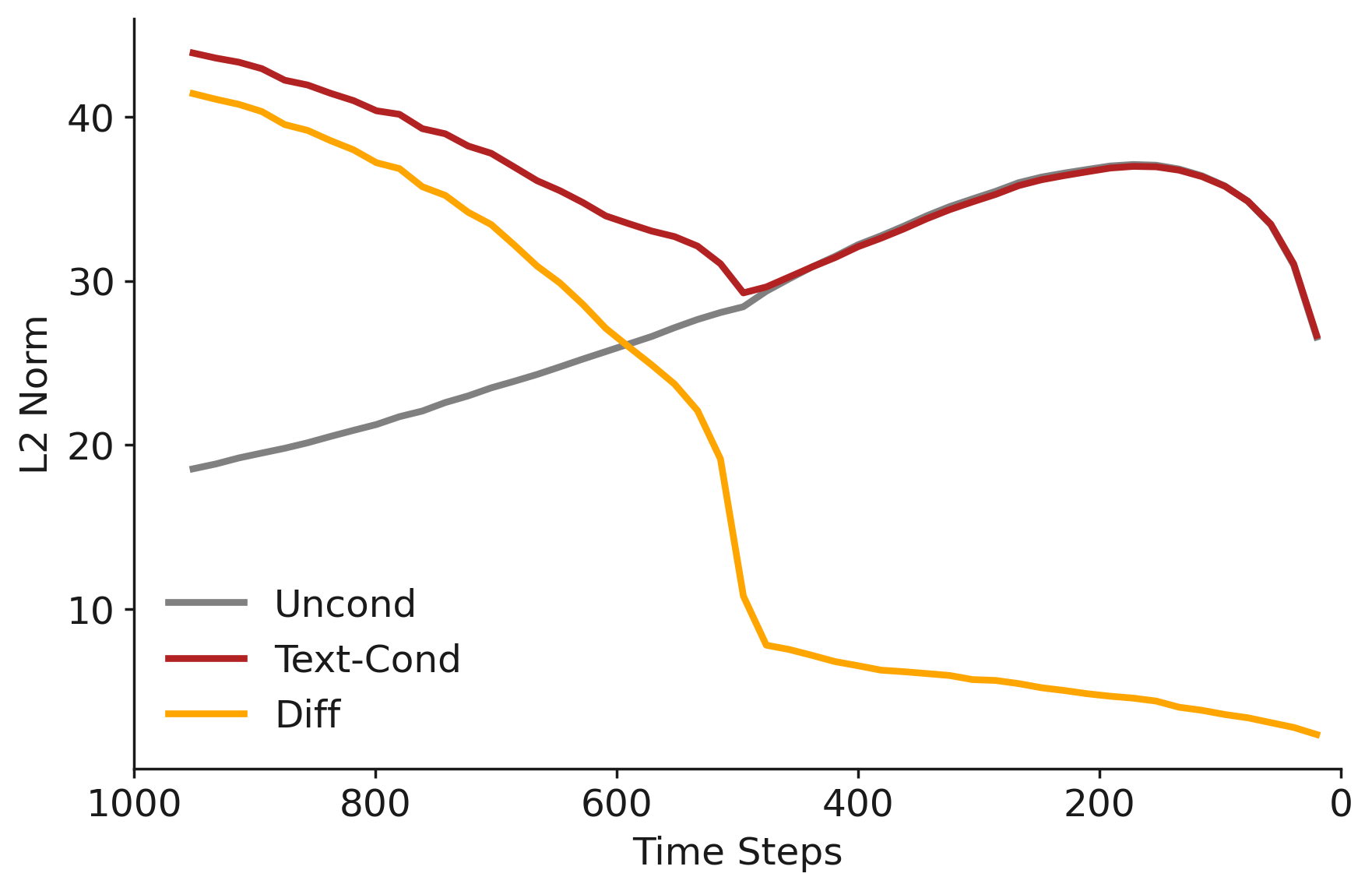}
    \caption{Average magnitude of the text-conditioned ($\epsilon_{\theta}(x_t, e_{p})$) and unconditional noise predictions ($\epsilon_{\theta}(x_t, e_{\emptyset})$) and their difference ($\epsilon_{\theta}(x_t, e_{p}) - \epsilon_{\theta}(x_t, e_{\emptyset})$) when applying zero CFG. We see a static transition point ($t=500$) appear when SDv2.1 is finetuned on the LAION-10k dataset \cite{somepalli2023understanding_neurips}.}
    \label{fig:l2_norms_sdv2}
\end{figure}

\begin{figure}[tbh]
    \centering
    \resizebox{\columnwidth}{!}{
    \begin{subfigure}[t]{0.11\textwidth}
        \centering
        \includegraphics[height=4.0in]{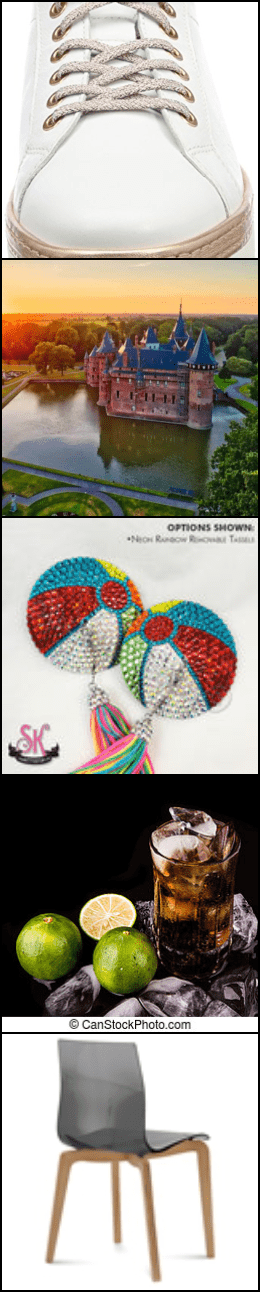}\\
        Training
    \end{subfigure}
    ~ 
    \begin{subfigure}[t]{0.11\textwidth}
        \centering
        \includegraphics[height=4.0in]{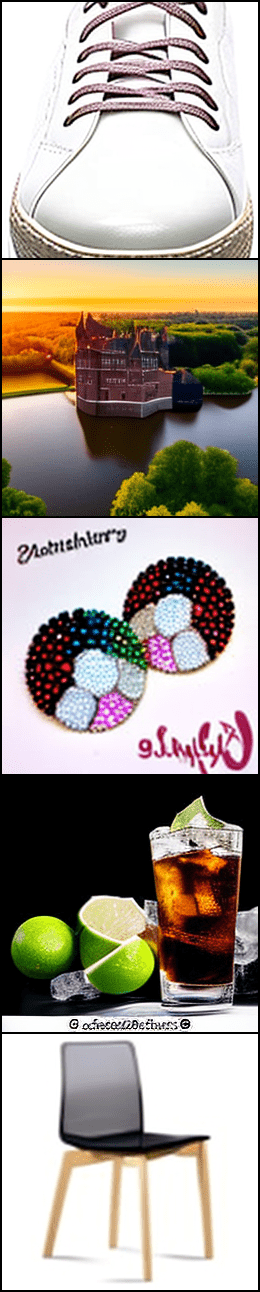}\\
        \lbrack1000, 0\rbrack
    \end{subfigure}
    ~
    \begin{subfigure}[t]{0.11\textwidth}
        \centering
        \includegraphics[height=4.0in]{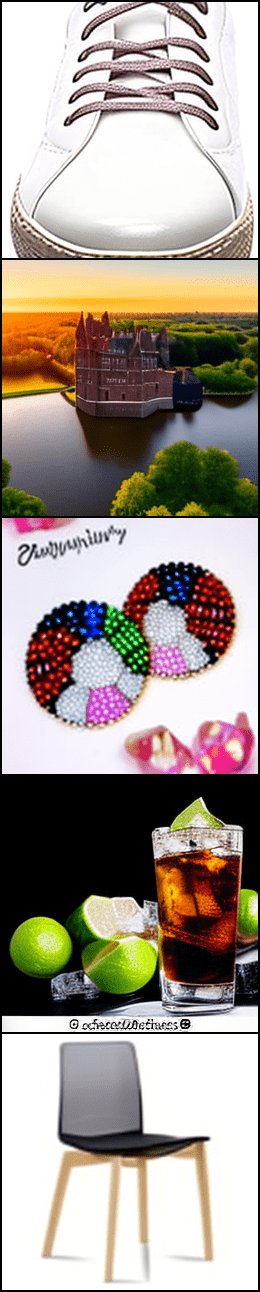}\\
        \lbrack800, 0\rbrack
    \end{subfigure}
    ~
    \begin{subfigure}[t]{0.11\textwidth}
        \centering
        \includegraphics[height=4.0in]{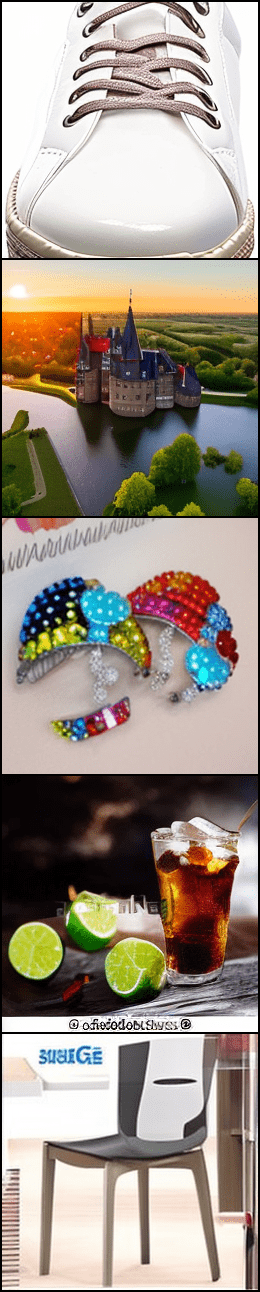}\\
        \lbrack600, 0\rbrack
    \end{subfigure}
    ~
    \begin{subfigure}[t]{0.11\textwidth}
        \centering
        \fcolorbox{green}{green}{        \hspace{-0.07in}\includegraphics[height=4.0in]{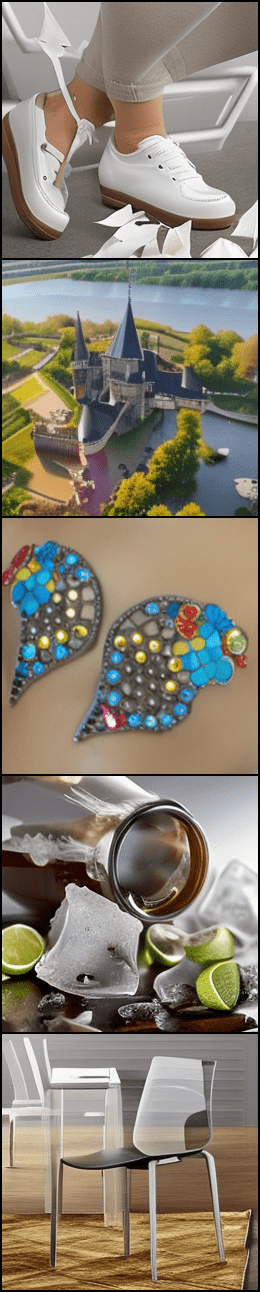}}\\
        \lbrack500, 0\rbrack
    \end{subfigure}
    ~
    \begin{subfigure}[t]{0.11\textwidth}
        \centering
        \includegraphics[height=4.0in]{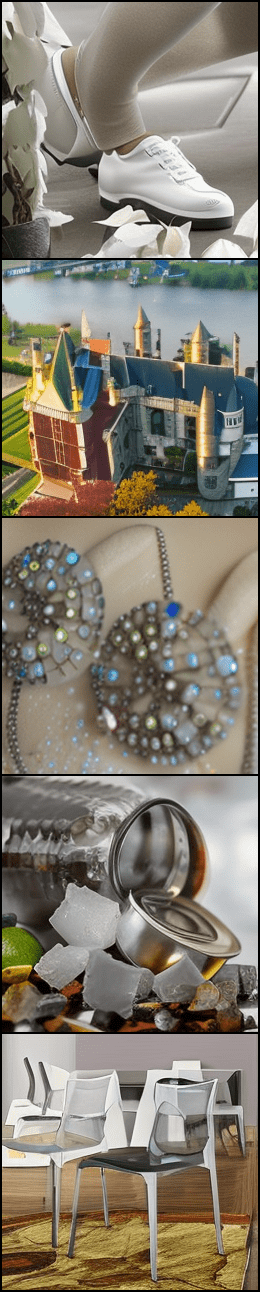}\\
        \lbrack400, 0\rbrack
    \end{subfigure}
    ~
    \begin{subfigure}[t]{0.11\textwidth}
        \centering
        \includegraphics[height=4.0in]{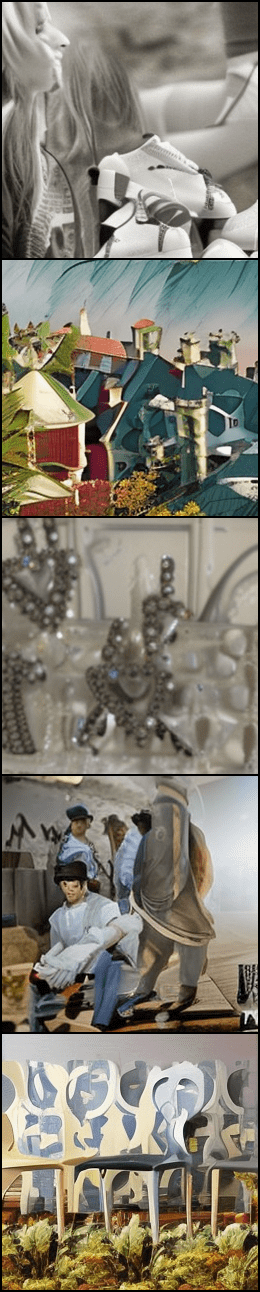}\\
        \lbrack200, 0\rbrack
    \end{subfigure}
    ~
    \begin{subfigure}[t]{0.11\textwidth}
        \centering
        \includegraphics[height=4.0in]{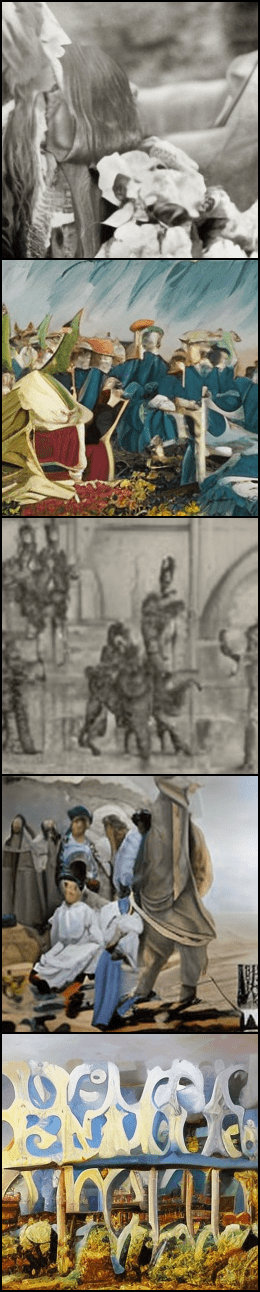}\\
        \lbrack0, 0\rbrack
    \end{subfigure}
    }
    \caption{Applying CFG before the static transition point (T=500) leads to memorized outputs while applying CFG after the fixed transition point leads to non-memorized outputs. Applying CFG too late results in poor-quality images that resemble the unconditional generations.}
    \label{fig:different_time_steps_laion10k}
\end{figure}

\paragraph{Static Transition Point.}

In some trained diffusion models, we observed a universal trend in the mean conditional $\epsilon_{\theta}(x_t, e_p)$ and unconditional noise prediction $\epsilon_{\theta}(x_t, e_{\emptyset})$ across different samples. As shown in Figure \ref{fig:l2_norms_sdv2}, the mean conditional noise prediction $\epsilon_{\theta}(x_t, e_p)$ is higher than unconditional noise prediction $\epsilon_{\theta}(x_t, e_{\emptyset})$ in the initial time steps ($t\geq 500$). At a static time step ($t=500$), the two scores align in magnitude until the end of the denoising process, and at the same time, the $L_2$ distance between them drastically drops. This suggests that the zero CFG trajectory leaves the attraction basin at that fixed time step ($t=500$). This can be further validated by applying CFG from any time step before the fixed transition point ($t=500$), as illustrated in Figure \ref{fig:different_time_steps_laion10k}. Applying CFG from an early time step leads to the same memorized image, but the output immediately switches at the transition point ($t=500$) such that we generate text-aligned non-memorized images. On the other hand, if we start applying CFG very late in the denoising process, the outputs are not well aligned with the text prompts and resemble their unconditional generations. 

\begin{figure}[]
    \includegraphics[width=\linewidth]{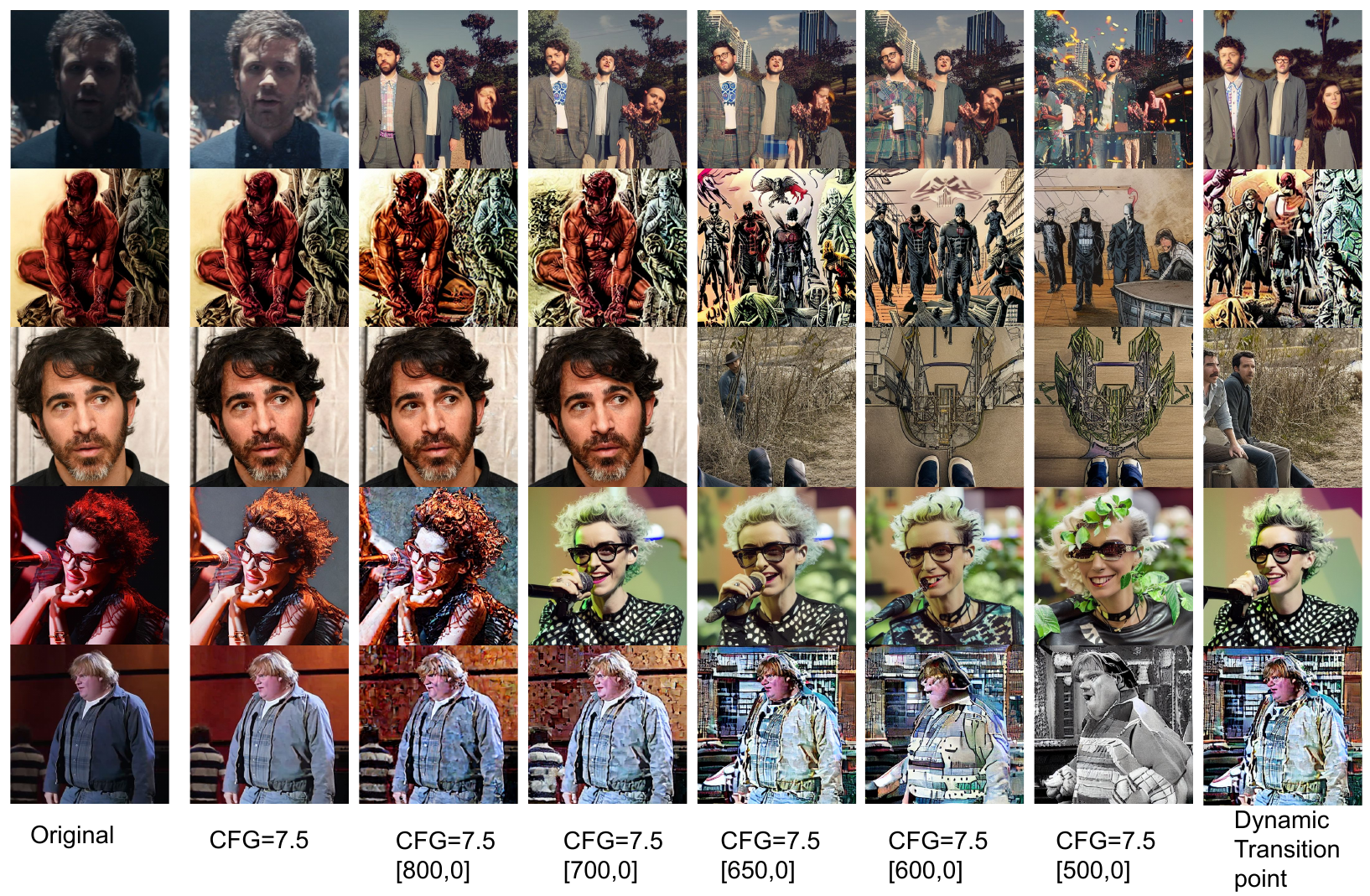}
    \caption{In some models, transition points can occur at a different time step for each sample, as seen for pre-trained SDv1.4. In row 1 the transition point is approximately $t=800$ while for row 2 it is $t=650$.}
    \label{fig:500_webster_dynamic}
\end{figure}

\paragraph{Dynamic Transition Point.}

\begin{figure}[htp]
    \centering
    \begin{subfigure}[t]{0.7\columnwidth}
        \centering
        \includegraphics[width=\linewidth]{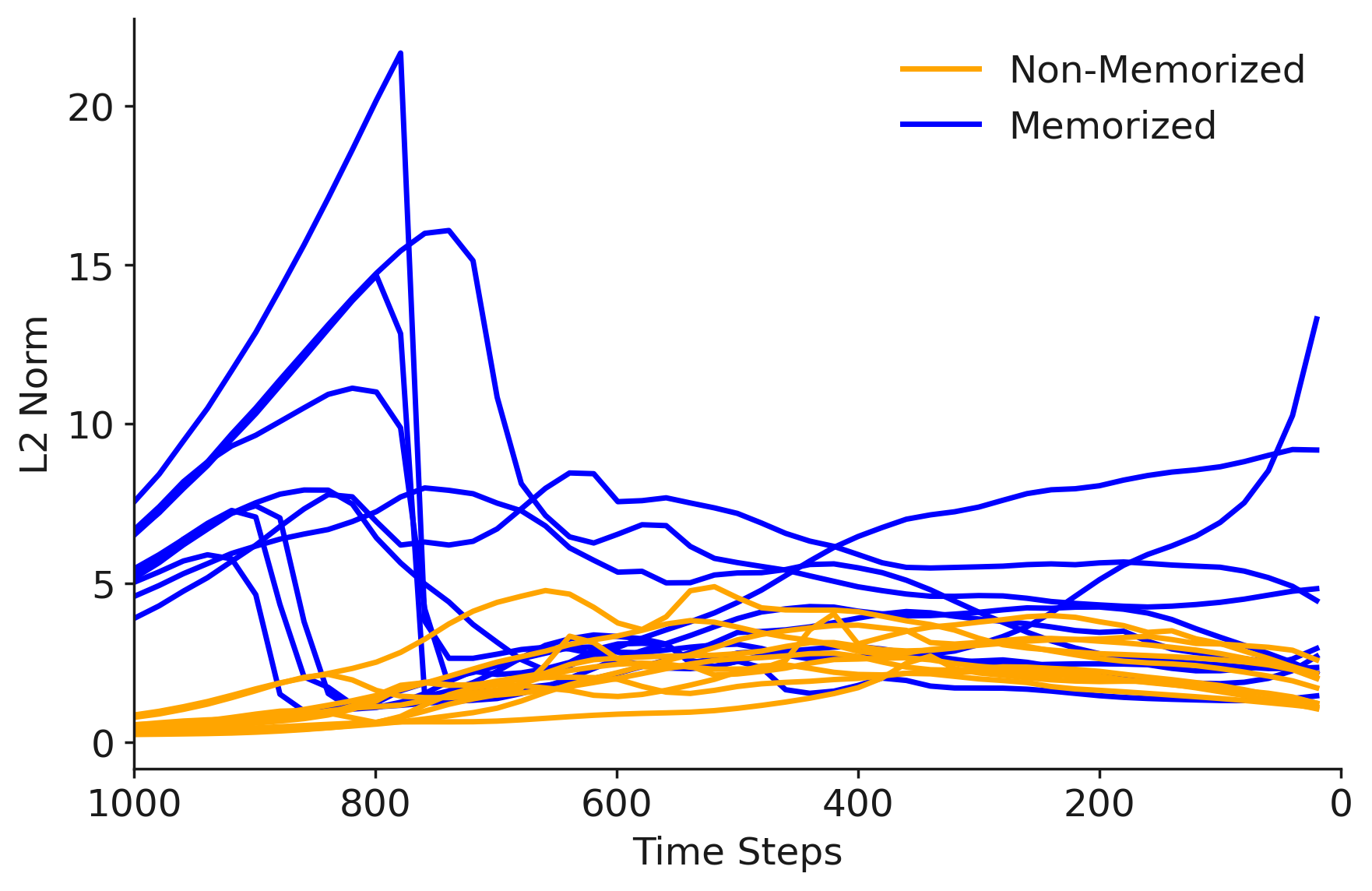}
        \caption{Applying zero CFG}
    \end{subfigure}
    ~
    \begin{subfigure}[t]{0.7\columnwidth}
        \centering
        \includegraphics[width=\linewidth]{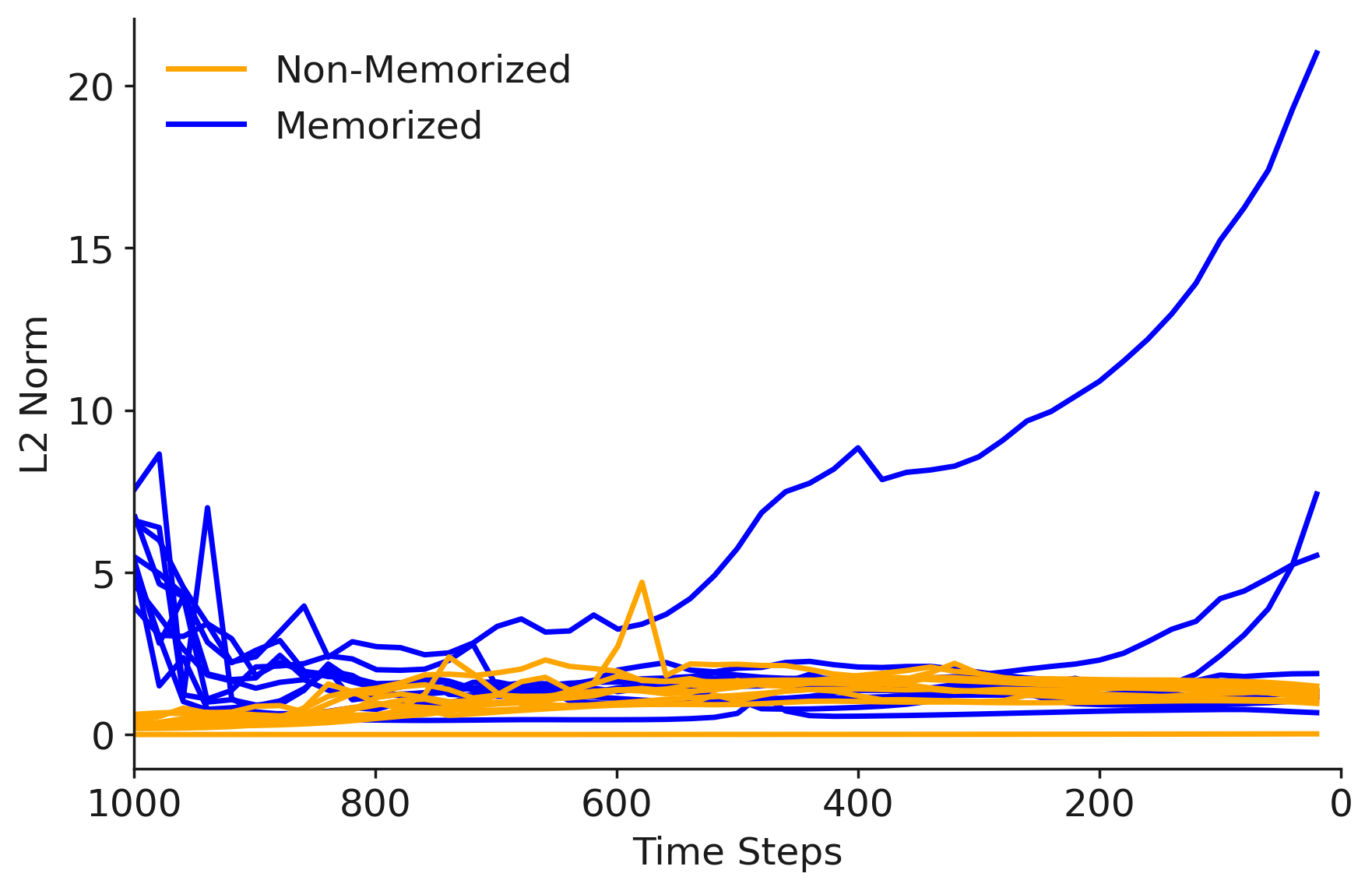}
        \caption{Applying opposite guidance }
    \end{subfigure}\\
    \caption{Magnitude of $\epsilon_{\theta}(x_t, e_p) - \epsilon_{\theta}(x_t, e_{\emptyset})$ when denoising (a) without CFG and (b) with opposite guidance. Transition point occurs sooner with opposite guidance.}
    \label{fig:dtpm_og_plots}
\end{figure}

In other trained diffusion models, we saw that every prompt and initialization pair leads to a different transition point. As shown in Figures \ref{fig:dtpm_og_plots}(a) and \ref{fig:500_webster_dynamic}, different trajectories leave the attraction basin at different time steps, characterized by their fall in the conditional guidance $\epsilon_{\theta}(x_t, e_p) - \epsilon_{\theta}(x_t, e_{\emptyset})$. We found that it is dependent on both the conditional embedding $e_p$ and the initialization $x_T$. We show an example in Figures \ref{fig:attraction_basin}(a) and \ref{fig:attraction_basin}(b), where the transition point occurs at different time steps even for the same prompt. Thus, we propose a method to find this dynamic transition point (DTP). As shown in Figure \ref{fig:attraction_basin}, it is the point after the first local minima in the graph. Thus, to mitigate memorization, this difference is tracked until the first local minimum occurs, and CFG is applied henceforth. This process requires no additional computations as the text-conditioned and unconditional noise predictions are computed by default when denoising using CFG. We present the pseudo-code for this approach in Algorithm \ref{alg:dcfg}.

    \begin{algorithm}[t]
    \small
           \caption{Reverse Diffusion with Dynamic Transition Point Method and Opposite Guidance}
           \label{alg:dcfg}
            \begin{algorithmic}[1]
             \Require $x_T \sim \mathcal{N}(0, \mathrm{I}_d), \lambda > 0, \upsilon_\text{OG} \in \{0,1\}$
             \State{$d_{T+2} = -\infty; d_{T+1} = -\infty$}
             \State{$s = -\lambda * \upsilon_\text{OG}$}
             \For{$t=T$ {\bfseries to} $1$}
                \State{$d_{t} = \left\|\epsilon_{\theta}(x_t, e_p) - \epsilon_{\theta}(x_t, e_{\emptyset})\right\|_2^2$}
                
                \If{$d_{t+2} > d_{t+1}$ {\bfseries and} $d_{t+1} < d_{t}$}
                \State{$s = \lambda $}
                \EndIf
                \State{$\hat{\epsilon} = \epsilon_{\theta}(x_t, e_{\emptyset}) + s (\epsilon_{\theta}(x_t, e_p) - \epsilon_{\theta}(x_t, e_{\emptyset}))$}
                 \State{$x_{t-1} =\frac{1}{\sqrt{\alpha}_t}(x_t - \frac{1- \alpha_t}{\sqrt{1-\Bar{\alpha_t}}}\hat{\epsilon})$}
              \EndFor
              \State {\bfseries return} $x_0$
            \end{algorithmic}
           
    \end{algorithm}
\subsection{Opposite Guidance}

An issue with this method is that when the transition point occurs very late $(t\leq500)$ in the denoising process, CFG is applied for fewer time steps, and this can impact the image quality. 

To ensure that the transition point occurs earlier, we introduce a new concept of opposite guidance (OG). Unlike negative prompting, we apply negative or opposite CFG, to push the diffusion trajectory in the opposite direction to that of traditional CFG and thus, push the trajectory away from the attraction basin of the traditional CFG trajectory sooner in the denoising process, as also shown in Figure \ref{fig:dtpm_og_plots}(b). Opposite guidance can be expressed as follows for $s>0$,

\begin{equation}
    \hat{\epsilon} \leftarrow \epsilon_{\theta}(x_t, e_{\emptyset}) - s (\underbrace{\epsilon_{\theta}(x_t, e_p) - \epsilon_{\theta}(x_t, e_{\emptyset})}_{\text{conditional guidance}}).
    \label{eq:negative_guidance}
\end{equation}

Similar to unconditional denoising, when denoising using opposite guidance, there is an ideal time step where the magnitude of conditional guidance falls. At this time step, switching from opposite guidance to traditional positive guidance leads to non-memorized samples with high image quality and text alignment. Thus, it can easily be integrated along with the previous method to find a static or dynamic transition point. We present the pseudo-code for this approach in conjunction with the DTP method in Algorithm \ref{alg:dcfg} with the parameter $\upsilon_\text{OG}=1$.

\section{Memorization under Different Scenarios}

We study memorization under different scenarios and show how this simple approach of finding an ideal transition point to avoid the attraction basin can be applied to effectively mitigate memorization.

\begin{itemize}
    \item Scenario 1: Fine-tuned SDv2.1 on 10,000 LAION datapoints (followed by \cite{somepalli2023understanding_neurips}). Memorization occurs due to overfitting on the small dataset. 
    \item Scenario 2: Fine-tuned SDv2.1 on the Imagenette dataset \citep{Howard_Imagenette_2019} containing 10 ImageNet classes (followed by \cite{somepalli2023understanding_neurips}).
    Exact memorization is not always observed but similarity with the training dataset increases. 
    \item Scenario 3: Fine-tuned SDv1.4 on 200 prompts duplicated 200 times with an additional 120,000 prompts that were not duplicated (followed by \cite{wen2024detecting}). Memorization occurs due to data duplication.
    \item Scenario 4: 500 memorized prompts for pre-trained SDv1.4 found using membership inference attack \citep{webster2023reproducible,carlini2023extracting} (followed by \cite{ren2024unveiling}). The exact cause of memorization for each prompt is unknown, but the strong presence of "trigger tokens" is noticeable. These are in the form of nouns such as celebrity and movie names with little to no additional descriptions. Results are in Appendix \ref{sec:scenario2}.
\end{itemize}

Notably, different researchers have independently studied these different scenarios, but, as we show later, their approaches do not generalize to other scenarios as memorization occurs due to different causes in each scenario. Refer to Appendix \ref{appendix:implementation_details} for other implementation details.

\paragraph{Evaluation Metrics} We utilize the similarity metric based on the SSCD \citep{pizzi2022self} embeddings to judge the level of memorization \citep{somepalli2023understanding_neurips}; the CLIP score~\citep{radford2021learning} to assess how well the generated images align with the text prompts; and the Fréchet Inception Distance (FID) \citep{heusel2017gans} for image quality.

\begin{figure*}[t!]
    \centering
    \resizebox{0.65\linewidth}{!}{
    \begin{subfigure}[t]{0.125\textwidth}
        \centering
        \includegraphics[height=5.4in]{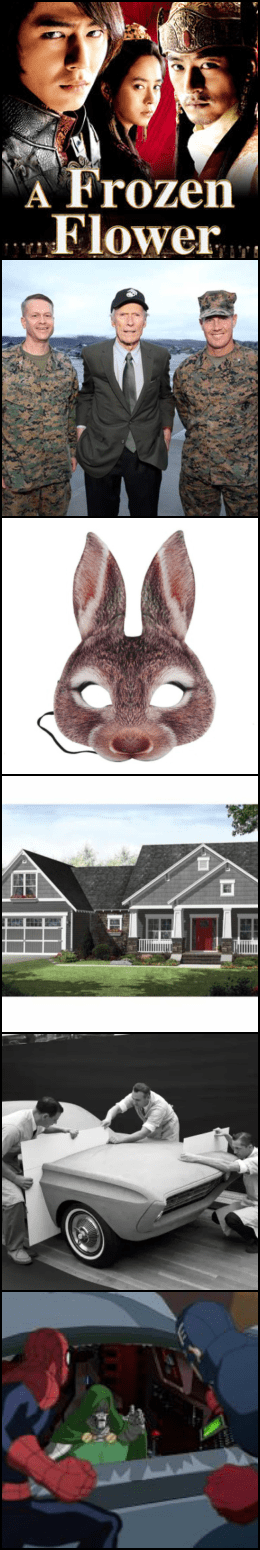}\\
        Training
    \end{subfigure}
    ~ 
    \begin{subfigure}[t]{0.125\textwidth}
        \centering
        \includegraphics[height=5.4in]{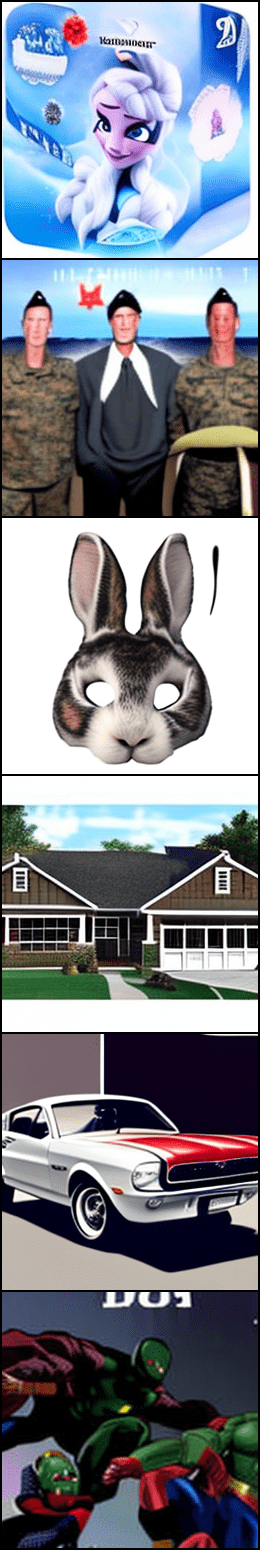}\\
        Add rand numb~\citep{somepalli2023understanding_neurips}
    \end{subfigure}
    ~
    \begin{subfigure}[t]{0.125\textwidth}
        \centering
        \includegraphics[height=5.4in]{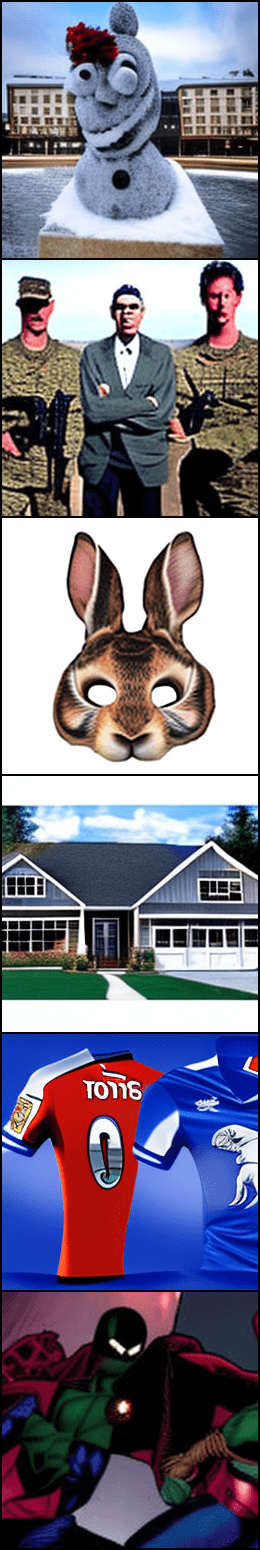}\\
        Add rand word~\citep{somepalli2023understanding_neurips}
    \end{subfigure}
    ~
    \begin{subfigure}[t]{0.125\textwidth}
        \centering
        \includegraphics[height=5.4in]{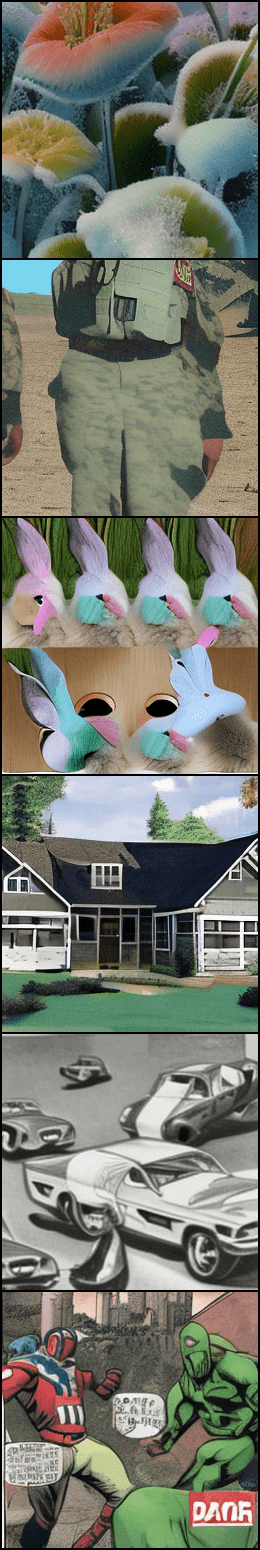}\\
        Wen et al.~\cite{wen2024detecting}
    \end{subfigure}
    ~
    \begin{subfigure}[t]{0.125\textwidth}
        \centering
        \includegraphics[height=5.4in]{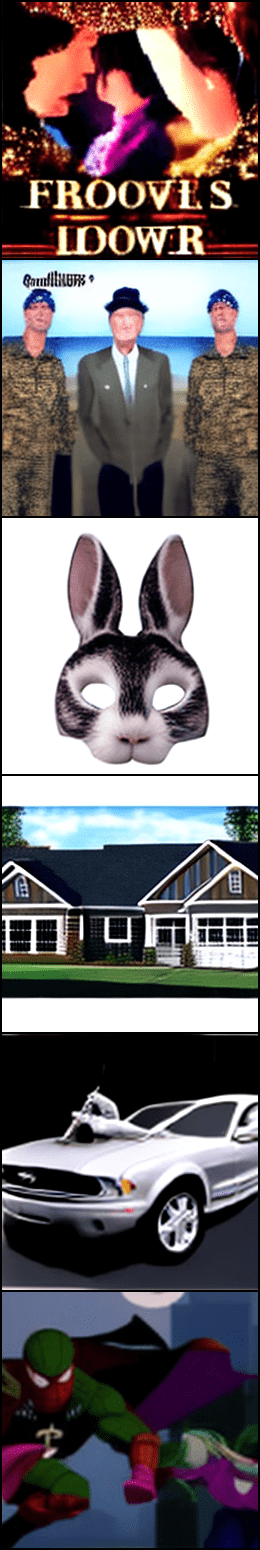}\\
        Ren et al.~\cite{ren2024unveiling}
    \end{subfigure}
    ~
    \begin{subfigure}[t]{0.125\textwidth}
        \centering
        \includegraphics[height=5.4in]{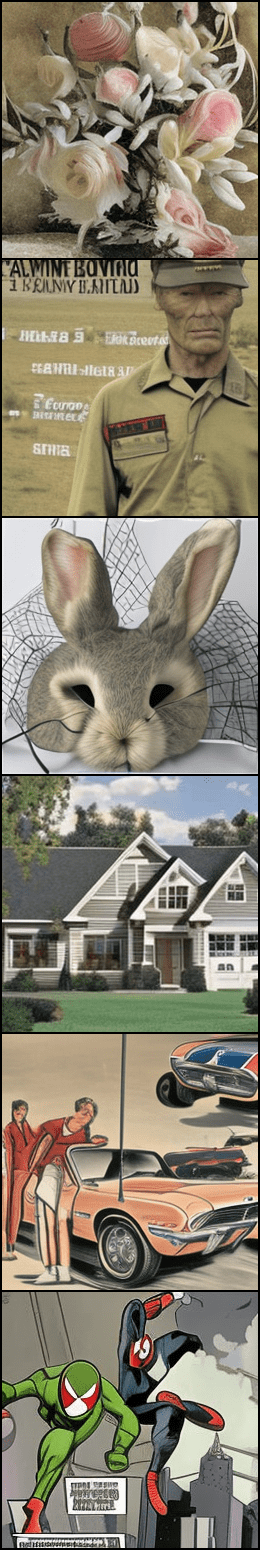}\\
        Ours (STP)
    \end{subfigure}
    ~
    \begin{subfigure}[t]{0.125\textwidth}
        \centering
        \includegraphics[height=5.4in]{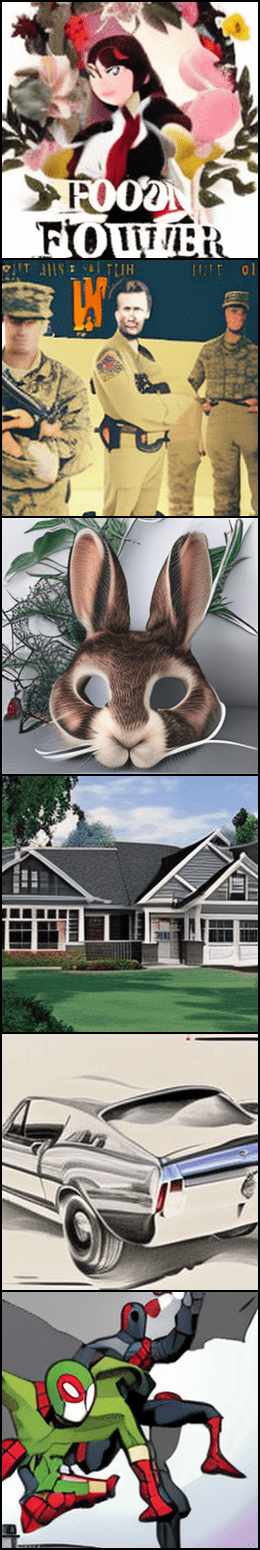}\\
        Ours (OG+STP)
    \end{subfigure}
    }
    \caption{Qualitative results comparing the proposed approach with the baselines in Scenario 1. Ren et al. \cite{ren2024unveiling} is unable to mitigate memorization in most cases while Wen et al. \cite{wen2024detecting} has poor image quality and text-alignment. More enlarged examples (Fig. \ref{fig:qual_comp_laion10k_appendix}) and prompts for this figure are in Appendix \ref{sec:prompts_figure_laion10k}.}
    \label{fig:qual_comp_laion10k}
\end{figure*}

\subsection{Scenario 1}

We observed that in this scenario, a static transition point at $t=500$ occurs as shown in Figure \ref{fig:l2_norms_sdv2}. We present qualitative and quantitative results in Figure \ref{fig:qual_comp_laion10k} and Table \ref{tab:laion-10k}, respectively. Our approach outperformed all previous methods, and we observed the previously proposed techniques that focused solely on the presence of "trigger tokens" \citep{wen2024detecting,ren2024unveiling} did not work well. This shows that when memorization occurs due to factors other than data duplication or the presence of trigger tokens, these methods may fail. We present further results when finetuning on the LAION-100k  dataset \cite{somepalli2023understanding_neurips} in Appendix \ref{sec:scenario3_additional_results}.

\begin{table}[]
    \centering
        \caption{Comparison of results on Scenario 1. None of the other approaches perform well. The only comparable method \cite{wen2024detecting} qualitatively results in poorer quality images (see visual examples in Fig. \ref{fig:qual_comp_laion10k}).}
        \label{tab:laion-10k}
        \resizebox{\columnwidth}{!}{
\begin{tabular}{lccc}
\hline 
                                    & \textbf{Similarity (95pc)} & \textbf{CLIP Score} & \textbf{FID} \\ \hline 
No Mitigation                            & 0.6504                       & 0.3027                        & 16.8373                 \\ \hline

Add rand word  \cite{somepalli2023understanding_neurips}                     & 0.5254                        & 0.2941                        & 17.4142                 \\
Add rand numb \cite{somepalli2023understanding_neurips}                   & 0.5416                        & 0.2993                        & 17.0245                 \\ \hline 
Wen et al. \cite{wen2024detecting}                       & 0.3853                         & 0.2895                        & 16.7176                \\ \hline 
Ren et al. \cite{ren2024unveiling}                          & 0.6028                        & 0.2959                         & 20.2931              \\ \hline 
Ours (STP)               & \textbf{0.2857}               & 0.2976                         & 19.8494                 \\
Ours (OG + STP) &  \underline{0.3811} & \textbf{0.3020} & \textbf{15.6679} \\ \hline
    \end{tabular}}
\end{table}

\subsection{Scenario 2} 

Even in this scenario, as illustrated in Figure \ref{fig:l2_norms_sdv2_imagenette}, a static transition point occurs. We applied CFG from $t=700$ and $t=600$ to show that we can reduce the similarity with the training dataset at no cost to the image quality, as shown in Table \ref{tab:imagenet}. We would like to point out that previous approaches are specific to text-to-image models and thus cannot be extrapolated to other conditioning mechanisms. This shows a further advantage of our CFG-based mitigation technique.

\begin{table}[htp]
\centering
\caption{Results on Scenario 2. The value in the brackets in the time steps between which we apply CFG. Applying CFG later from $t=700$ (row 2) and $t=600$ (row 3) reduces similarity with comparable or improved FID score.}
\label{tab:imagenet}
\begin{tabular}{lcc}
\hline
                    & \textbf{Similarity (95pc)} & \textbf{FID}      \\ \hline
No mitigation     & 0.3702   & 43.35 \\ \hline
CFG=7.5 {[}700,0{]} & 0.3018   & \textbf{38.86} \\
CFG=7.5 {[}600,0{]} & \textbf{0.2756}    & 45.73 \\ \hline
\end{tabular}
\end{table}

\begin{figure*}[h!]
    \centering
    \resizebox{0.65\linewidth}{!}{
    \begin{subfigure}[t]{0.125\textwidth}
        \centering
        \includegraphics[height=5.5in]{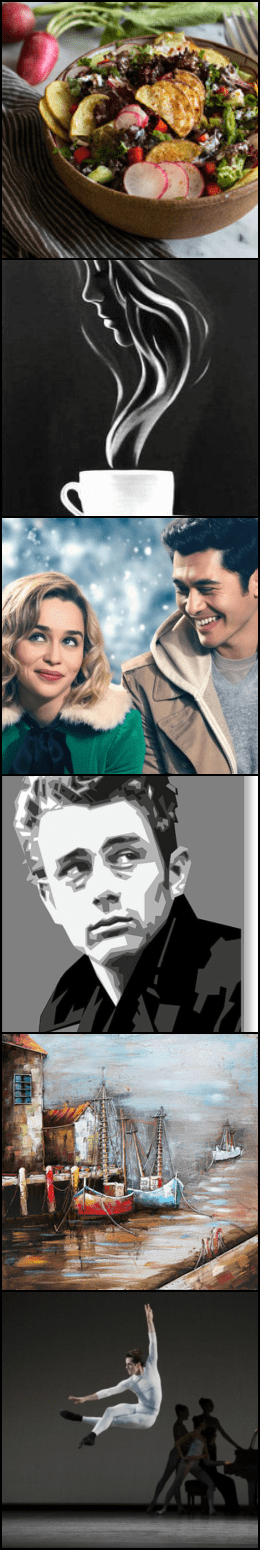}\\
        Training
    \end{subfigure}
    ~ 
    \begin{subfigure}[t]{0.125\textwidth}
        \centering
        \includegraphics[height=5.5in]{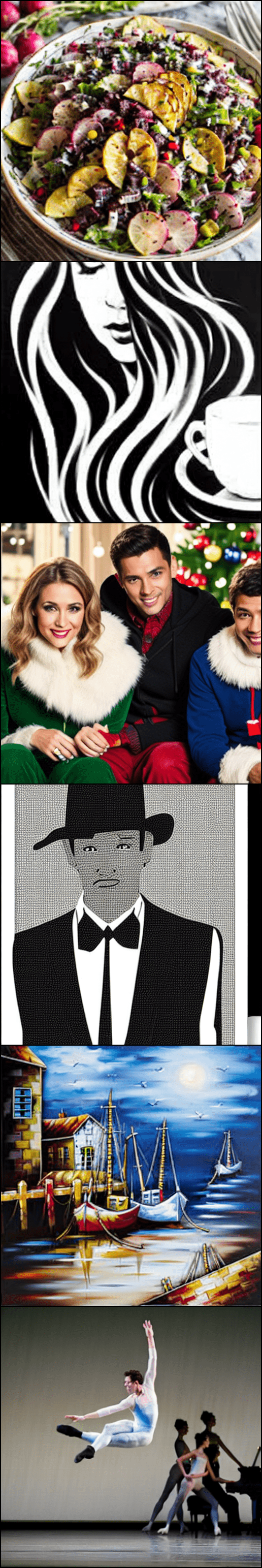}\\
        Add rand numb~\citep{somepalli2023understanding_neurips}
    \end{subfigure}
    ~
    \begin{subfigure}[t]{0.125\textwidth}
        \centering
        \includegraphics[height=5.5in]{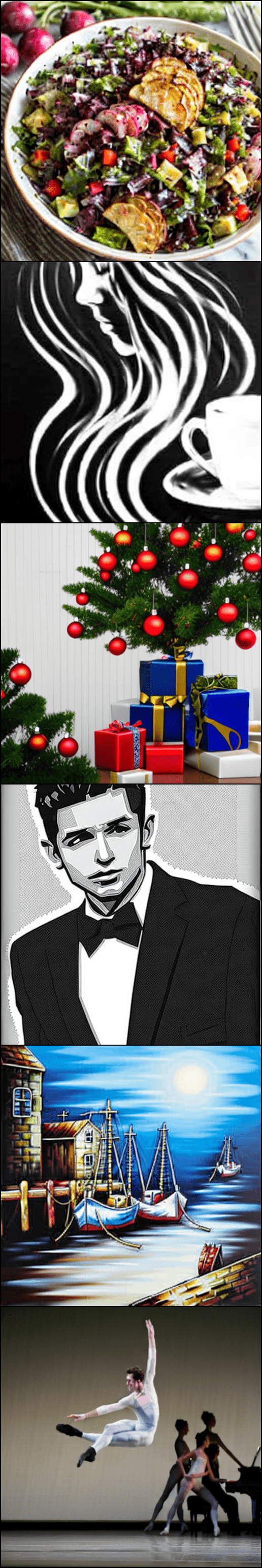}\\
        Add rand word~\citep{somepalli2023understanding_neurips}
    \end{subfigure}
    ~
    \begin{subfigure}[t]{0.125\textwidth}
        \centering
        \includegraphics[height=5.5in]{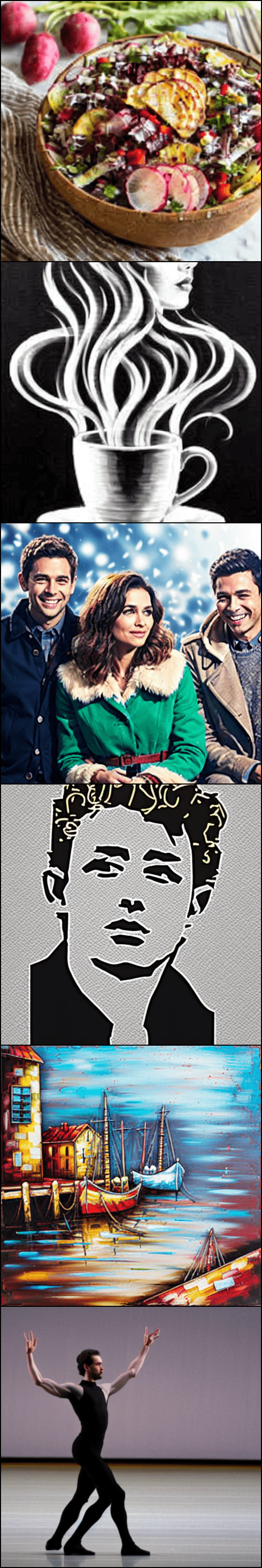}\\
        Wen et al.~\cite{wen2024detecting}
    \end{subfigure}
    ~
    \begin{subfigure}[t]{0.125\textwidth}
        \centering
        \includegraphics[height=5.5in]{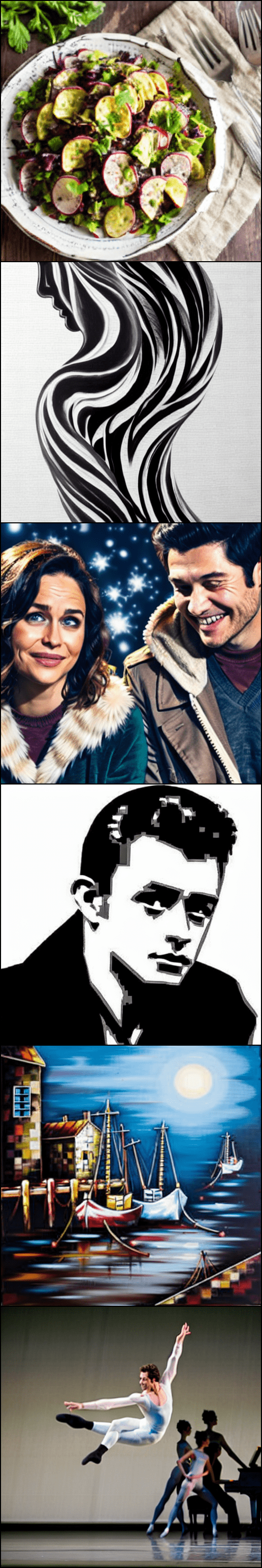}\\
        Ren et al.~\cite{ren2024unveiling}
    \end{subfigure}
    ~
    \begin{subfigure}[t]{0.125\textwidth}
        \centering
        \includegraphics[height=5.5in]{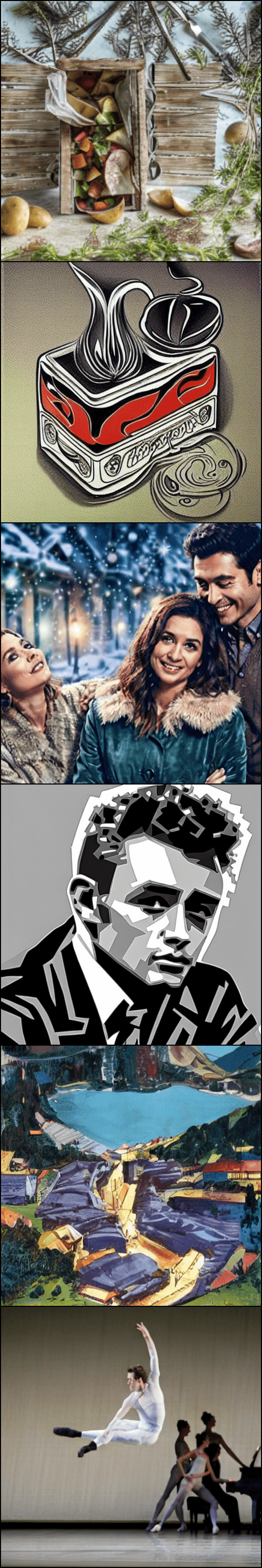}\\
        Ours (DTP)
    \end{subfigure}
    ~
    \begin{subfigure}[t]{0.125\textwidth}
        \centering
        \includegraphics[height=5.5in]{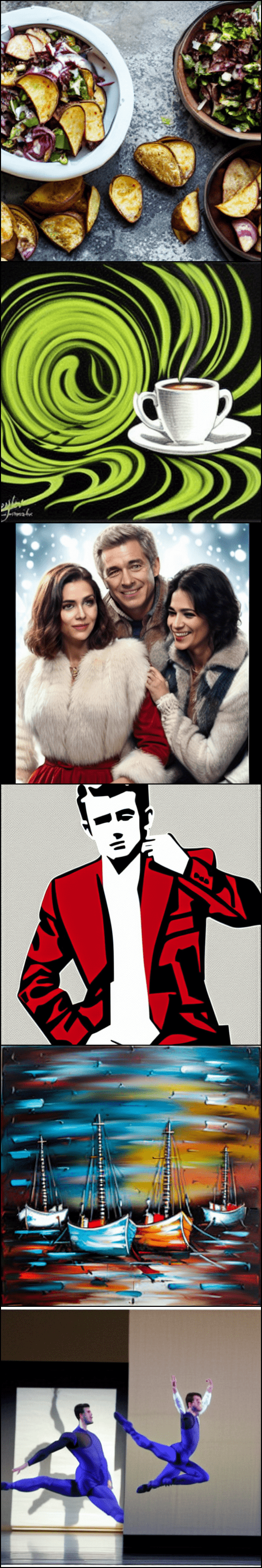}\\
        Ours (OG+DTP)
    \end{subfigure}
    }
    \caption{Qualitative results comparing the proposed approach with the baselines in Scenario 3. Prompts used to generate these figures are given in the Appendix. Rows 1, 2, 5, and 6 show examples, where for most baselines the output images remain closely related to their memorized versions. More enlarged examples (Fig. \ref{fig:qual_comp_200_memorized_images_appendix}) and prompts for this figure are in Appendix \ref{sec:prompts_figure_200_memorized}.}
    \label{fig:qual_comp_200_memorized_images}
\end{figure*}

\subsection{Scenario 3}

\begin{table}[]
    \centering
        \caption{Table showcasing results in Scenario 3. Wen et al. \cite{wen2024detecting} studied this scenario and is the only other method which performs well but does not generalize to other scenarios (see Table \ref{tab:laion-10k}). }
        \label{tab:200_memorized_images_dup}
        \resizebox{\columnwidth}{!}{
\begin{tabular}{lcccc}
\hline 
 & \multicolumn{2}{c}{\textbf{Similarity}} & \multirow{2}{*}{\textbf{CLIP Score}} & \multirow{2}{*}{\textbf{FID}} \\
 & \textbf{95pc} & \textbf{Mean $\pm$ Std} & & \\ \hline 
No Mitigation                       & 0.7977  & 0.5513 $\pm$ 0.16  & 0.3105    & 106.49 \\ \hline
Add rand word   \cite{somepalli2023understanding_neurips}              & 0.7480    & 0.4312 $\pm$ 0.21  & 0.3071     & 116.92 \\
Add rand numb \cite{somepalli2023understanding_neurips}                & 0.7366  & 0.3850 $\pm$ 0.21  & 0.3027    & 126.54  \\ \hline 
Wen et al. \cite{wen2024detecting} ($l_{target}$=3)             & 0.7747   & 0.5147 $\pm$ 0.17  & \textbf{0.3100}     & \textbf{109.28} \\ \hline 
Wen et al. \cite{wen2024detecting} ($l_{target}$=1)        & \underline{0.6038}  & \textbf{0.2808 $\pm$ 0.16}  & 0.3050    & 136.34 \\ \hline 
Ren et al. \cite{ren2024unveiling}                      & 0.6881  & 0.4036 $\pm$ 0.16  & 0.3066     & 124.38 \\ \hline 
Ours (DTP) & \textbf{0.5885}   & \textbf{0.2866 $\pm$ 0.16}  & 0.3020      & 138.92 \\ 
Ours (OG + DTP) & 0.6915  & \textbf{0.2844 $\pm$ 0.20}  &  0.2910 & 140.05 \\ \hline 
\end{tabular}}
\end{table}

In this scenario, we observed that a dynamic transition point occurs. We experimentally validate our approach of DTP method and opposite guidance qualitatively and quantitatively in Figure \ref{fig:qual_comp_200_memorized_images} and Table \ref{tab:200_memorized_images_dup}. We saw that our method performs at par with \cite{wen2024detecting} that studied this specific scenario and even outperforms it in terms of memorization in the top 95 percentile. Approaches by Ren et al. \cite{ren2024unveiling} and Somepalli et al.\cite{somepalli2023understanding_neurips} that were proposed for other scenarios do not work well, showcasing their lack of generalizability.

\paragraph{Inference Time.} One of the major advantages of our approach is its simplicity and lack of computational time overhead. Image generation using our approach as well as standard CFG-based image generation using SDv2.1 on one A100 GPU takes 1.26 seconds, while in comparison the mitigation technique proposed by Wen et al. \cite{wen2024detecting} took 2.86 seconds and that of Ren et al. \cite{ren2024unveiling} took 2.08 seconds. These are the average values across 10,000 generations.

\paragraph{Limitations.} As transition points do not exist for non-memorized samples, we require detecting memorization before we can apply our approach. This can either be done by denoising twice, once with traditional CFG to look at the trends in the magnitude of conditional guidance and the second time to apply our mitigation technique if the prompt is memorized. Alternatively, as \cite{wen2024detecting,ren2024unveiling} showed, we can do detection at $t=0$ by simply looking at the magnitude of conditional guidance $\| \epsilon_{\theta}(x_t, e_p) - \epsilon_{\theta}(x_t, e_{\emptyset}) \|_2^2$. If this magnitude is above a certain threshold, it indicates that the sample is memorized with high accuracy and we can apply our guidance accordingly.

\section{Conclusion}

We present a novel understanding of the dynamics behind memorization in conditional diffusion models. We showcase the presence of an attraction basin that steers randomly initialized latent vectors towards a memorized output. We propose an approach to steer away from the attraction basin and detect the point at which the trajectory leaves the attraction basin, referred to as the transition point. We show that applying CFG after the transition point leads to non-memorized outputs. We presented results in various scenarios such as the presence of data replication, fine-tuning on smaller datasets, and memorized examples found in existing pre-trained models to showcase its efficiency.  

In summary, our method is simple, works well across all tested scenarios in terms of both image quality metrics and visual inspection, requires less computational time, and does not require altering the prompt.

{
    \small
    \bibliographystyle{ieeenat_fullname}
    \bibliography{biblio}
}

\newpage


\clearpage
\setcounter{page}{1}
\maketitlesupplementary

\begin{figure}[btp]
    \centering
    \includegraphics[width=0.8\linewidth]{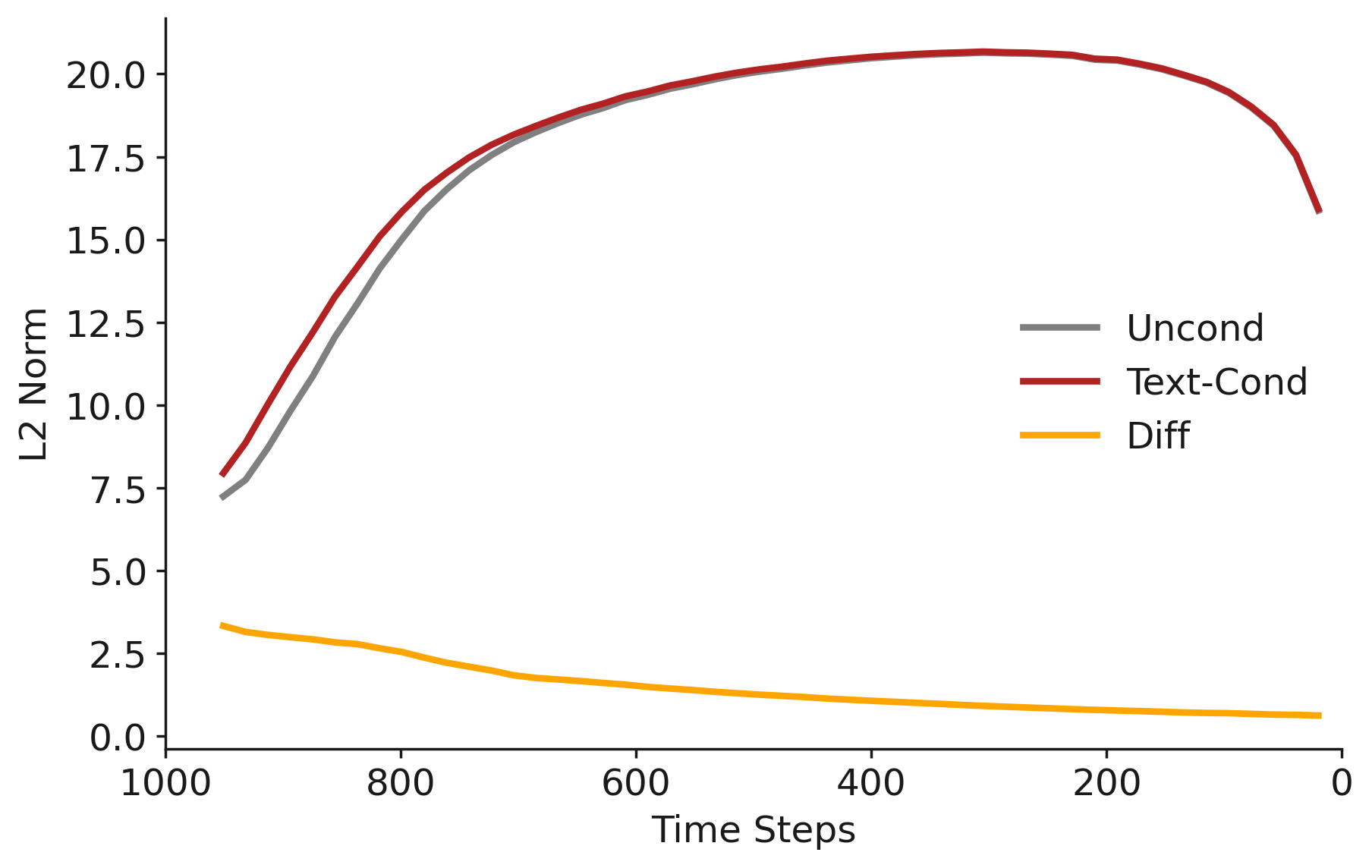}
    \caption{Mean conditional and unconditional noise predictions when SDv2.1 is finetuned on the Imagenette dataset.}
    \label{fig:l2_norms_sdv2_imagenette}
\end{figure}

We present the following contents in the Appendix:
\begin{itemize}
    \item Additional analysis on the attraction basin in Section \ref{sec:attraction_basin_extra_appendix}. 
    \item Experimental results on Scenario 4 that were discussed in the main text in Section \ref{sec:scenario2}. 
    \item Additional analysis of Scenario 1, where we provide experimental results when SDv2.1 is finetuned on LAION-100K dataset in Section \ref{sec:scenario3_additional_results}. 
    \item Figure on Scenario 2, showing the occurrence of a static transition point in Figure \ref{fig:l2_norms_sdv2_imagenette}. 
    \item Prompts used in Figure \ref{fig:qual_comp_laion10k} and Figure \ref{fig:qual_comp_200_memorized_images} in Sections \ref{sec:prompts_figure_laion10k} and \ref{sec:prompts_figure_200_memorized} respectively. 
    \item Additional examples of transition points coinciding with a fall in conditional guidance in Figure \ref{fig:chris_melissa}. 
    \item More visual results on different scenarios, comparing with baselines in Section \ref{sec:appendix_visual_examples}.
\end{itemize}

\section{Additional Analysis of the Attraction Basin}
\label{sec:attraction_basin_extra_appendix}

\begin{figure}[htp]
    \centering
    \includegraphics[width=\linewidth]{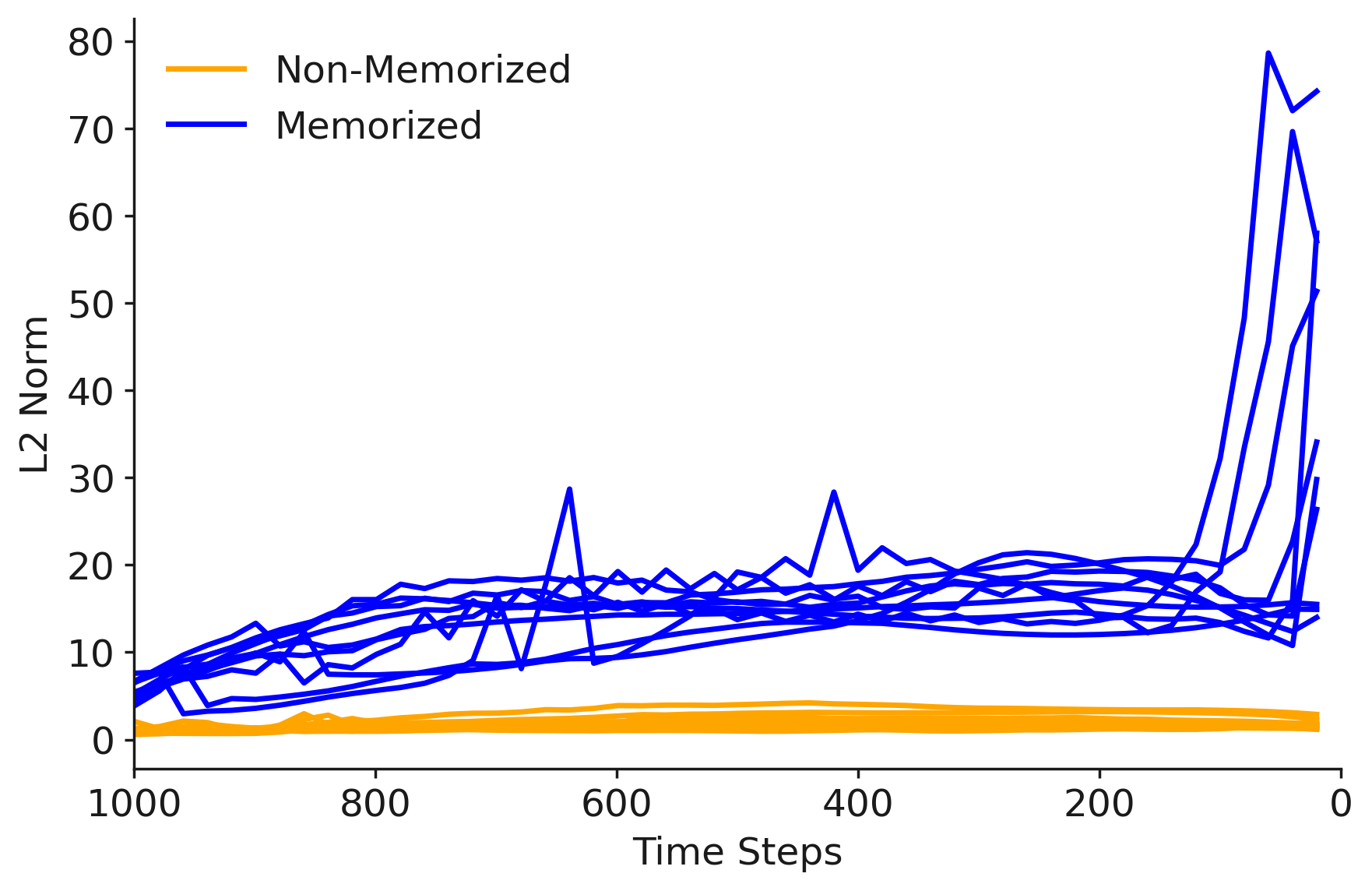}
    \caption{When you apply CFG from the beginning, the magnitude of $\epsilon_{\theta}(x_t, e_p) - \epsilon_{\theta}(x_t, e_{\emptyset})$ remains high throughout.}
    \label{fig:l2_norms_cfg}
\end{figure}

In the paper, we discuss observing the attraction basin when applying zero CFG in the denoising process by observing the magnitude of $\epsilon_{\theta}(x_t, e_p) - \epsilon_{\theta}(x_t, e_{\emptyset})$. Now the question arises, what is the trend when denoising with CFG? Does the value still drop after a particular time step? We show in Figure \ref{fig:l2_norms_cfg} that the magnitude of $\epsilon_{\theta}(x_t, e_p) - \epsilon_{\theta}(x_t, e_{\emptyset})$ remains high throughout the denoising process when we denoise using CFG. Further validating our observation, as in this case, the sample will be inside the attraction basin throughout the denoising process.

Interestingly, our observations also align with previous studies on improving diversity and fidelity in diffusion models where they showed the negative impacts of high CFG in the initial time steps \citep{wang2024analysis,kynkaanniemi2024applying,chang2023muse} by showing that monotonically increasing CFG weight schedulers lead to improved performance. These studies, however, are not in the context of memorization.

\begin{figure*}[btp]
    \centering
    \scalebox{1.0}{
    \begin{subfigure}[t]{0.125\textwidth}
        \centering
        \includegraphics[height=4.75in]{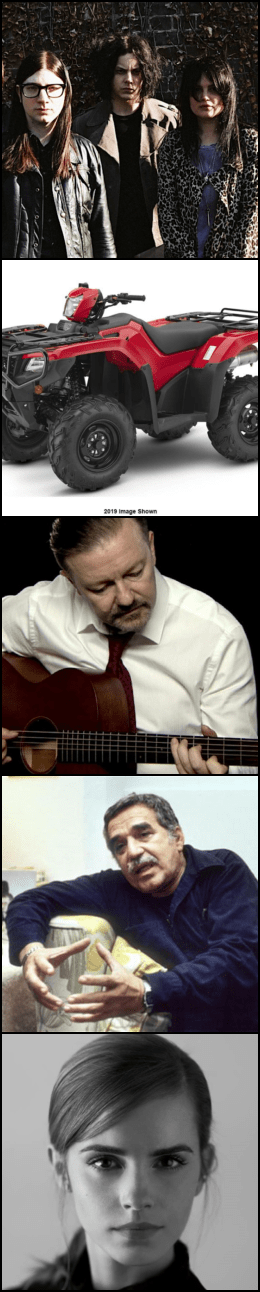}\\
        Training
    \end{subfigure}
    ~ 
    \begin{subfigure}[t]{0.125\textwidth}
        \centering
        \includegraphics[height=4.75in]{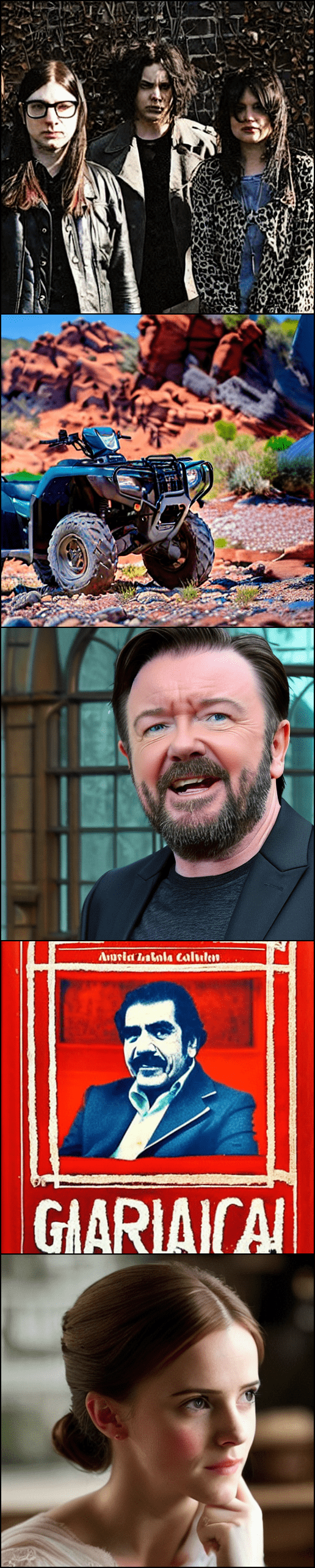}\\
        Add rand numb~\citep{somepalli2023understanding_neurips}
    \end{subfigure}
    ~
    \begin{subfigure}[t]{0.125\textwidth}
        \centering
        \includegraphics[height=4.75in]{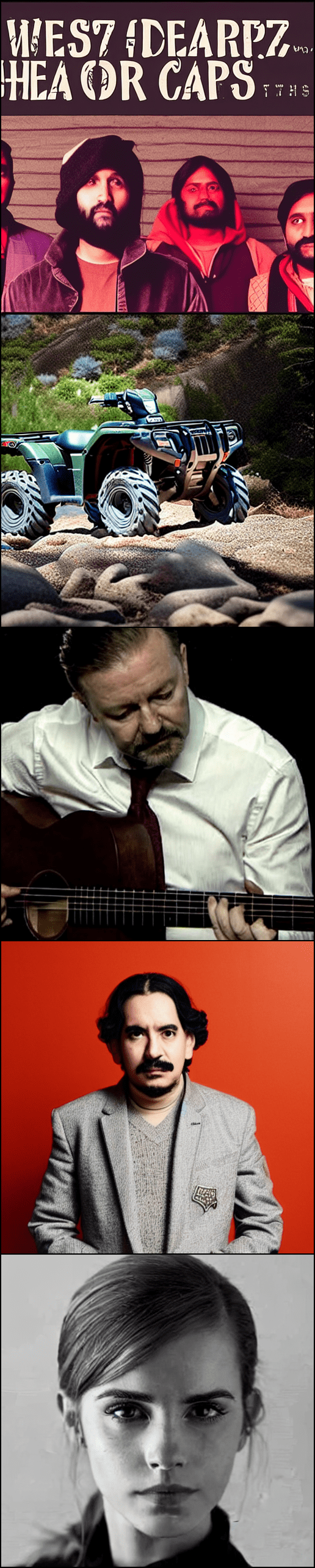}\\
        Add rand word~\citep{somepalli2023understanding_neurips}
    \end{subfigure}
    ~
    \begin{subfigure}[t]{0.125\textwidth}
        \centering
        \includegraphics[height=4.75in]{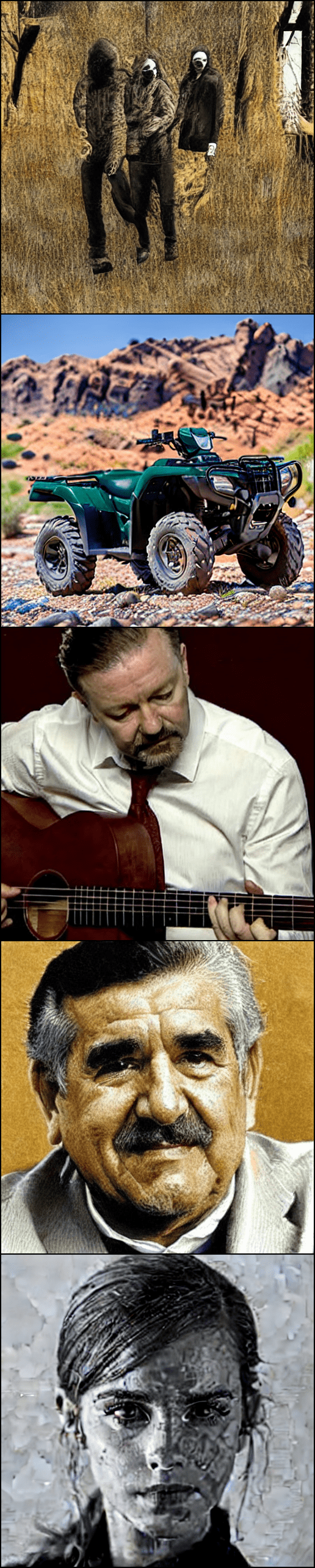}\\
        Wen et al.~\cite{wen2024detecting}
    \end{subfigure}
    ~
    \begin{subfigure}[t]{0.125\textwidth}
        \centering
        \includegraphics[height=4.75in]{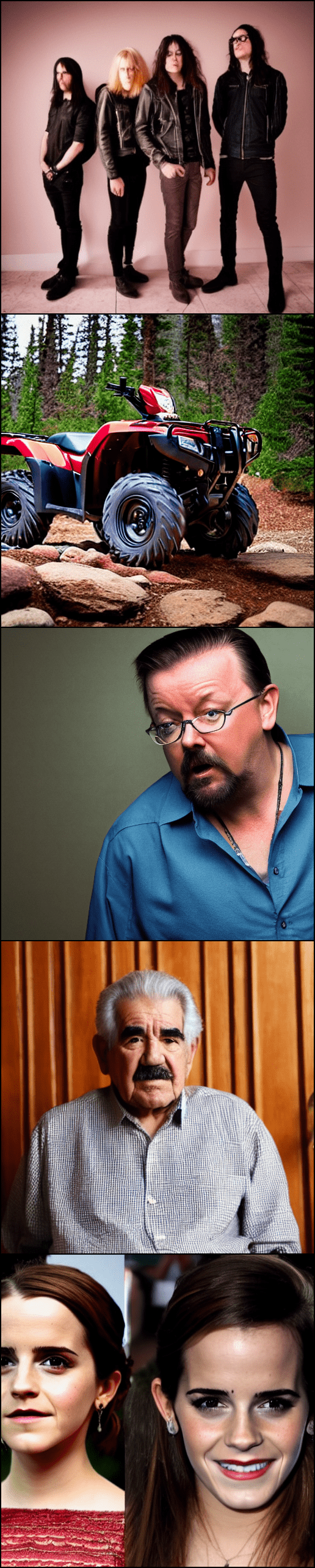}\\
        Ren et al.~\cite{ren2024unveiling}
    \end{subfigure}
    ~
    \begin{subfigure}[t]{0.125\textwidth}
        \centering
        \includegraphics[height=4.75in]{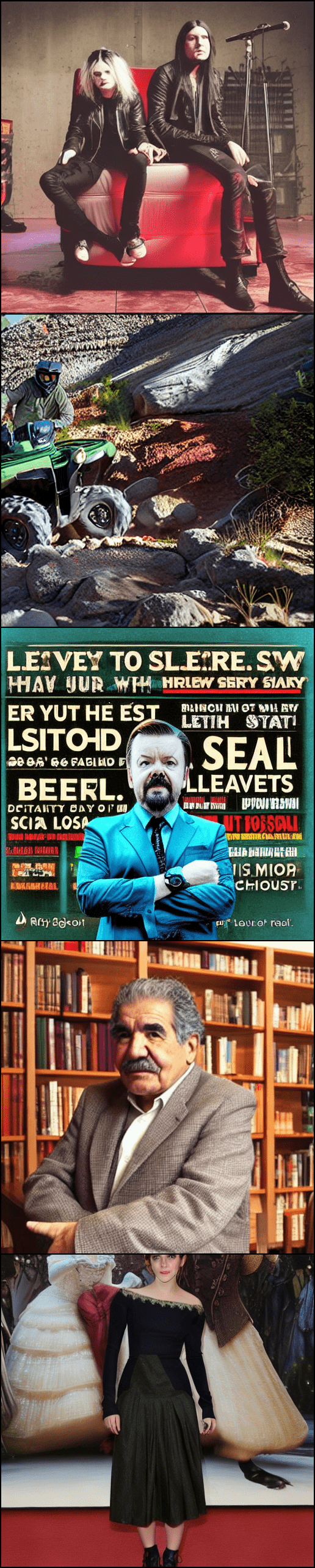}\\
        Ours (DTP)
    \end{subfigure}
    ~
    \begin{subfigure}[t]{0.125\textwidth}
        \centering
        \includegraphics[height=4.75in]{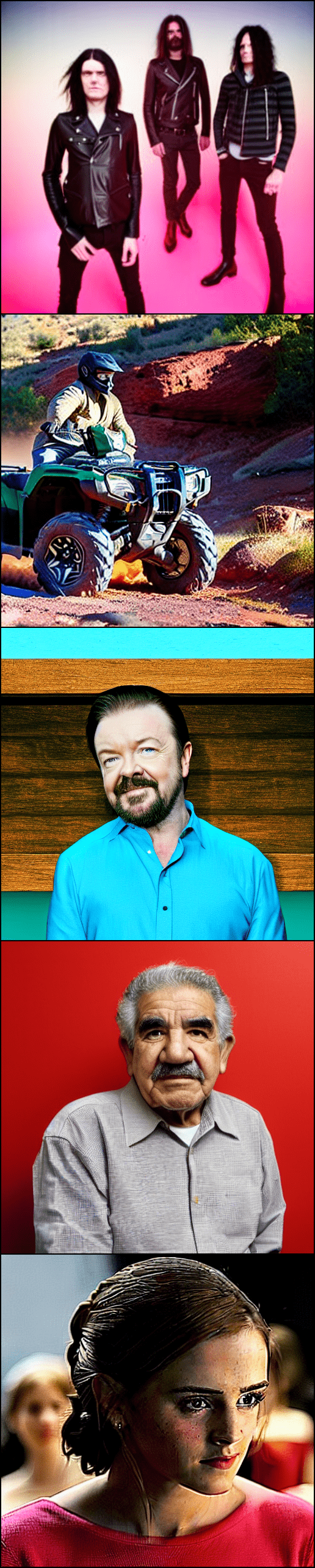}\\
        Ours (OG+DTP)
    \end{subfigure}
    }
    \caption{Qualitative results comparing the proposed approach with the baselines in Scenario 4. 
    Prompts -  (a) """Listen to The Dead Weather's New Song, """"Buzzkill(er)"""""""; (b) 2020 Honda FourTrax Foreman Rubicon 4x4 Automatic DCT in New Haven, Connecticut - Photo 1; (c) "Listen to Ricky Gervais Perform ""Slough"" as David Brent"; (d) Gabriel García Márquez's Collection Is Going to Austin; (e) Emma Watson to play Belle in Disney's <i>Beauty and the Beast</i>}
    \label{fig:qual_comp_500_webster}
\end{figure*}

\section{Experimental Results on Scenario 4}
\label{sec:scenario2}

For Scenario 4, where memorization occurs due to the presence of trigger words, we observed a dynamic transition point. We apply the same approach as present in Alg. \ref{alg:dcfg}.  

\paragraph{Experimental Results:} We compare our approach with previous baselines in Figure \ref{fig:qual_comp_500_webster} and Table \ref{tab:500_webster}, and show that our simple approach is able to mitigate memorization in this scenario as well. \cite{ren2024unveiling} had initially studied this scenario and we report comparable similarity results while still being generalizable to other scenarios. Our opposite guidance and dynamic transition point method yield a 0.2611 similarity score as compared to theirs of 0.2544. Additionally, we do not deteriorate the image quality as observed by the FID score of 155, which is the same as the FID score without any mitigation strategy.  Other approaches such as \cite{wen2024detecting} and \cite{somepalli2023understanding_neurips} lead to poorer similarity scores.

\begin{table}[]
    \centering
        \caption{Results on Scenario 4. Ren et al. \cite{ren2024unveiling} studied this scenario.}
        \label{tab:500_webster}
         \resizebox{\columnwidth}{!}{
\begin{tabular}{lcccc}
\hline 
 & \multicolumn{2}{c}{\textbf{Similarity}} & \multirow{2}{*}{\textbf{CLIP Score}} & \multirow{2}{*}{\textbf{FID}} \\
 & \textbf{95pc} & \textbf{Mean $\pm$ Std} & & \\ \hline 
No mitigation                                 & 0.9262               & 0.5508 $\pm$ 0.31          & 0.3144                                     & 155.35                                \\ \hline 
Add rand word \cite{somepalli2023understanding_neurips}  & 0.9049                 & 0.3779 $\pm$ 0.29          & 0.3046                                      & \textbf{143.10}                       \\
Add rand numb \cite{somepalli2023understanding_neurips} & 0.8934                & 0.3740 $\pm$ 0.28         & 0.3068                                   & 146.01                                \\ \hline
Wen et al. \cite{wen2024detecting} ($l_{target}$ =3)                & 0.9155              & 0.4913 $\pm$ 0.31          & \textbf{0.3134}                                & 154.70                               \\
Wen et al. \cite{wen2024detecting} ($l_{target}$ =1)                & 0.9011                  & 0.3287 $\pm$ 0.27        & 0.3088                                    & 147.74                                \\ \hline

Ren et al. \cite{ren2024unveiling}                                  & \textbf{0.6718}         & \textbf{0.2544 $\pm$ 0.18} & 0.3110                                        & 148.93                                \\ \hline 
Ours (DTP)                                         & 0.8722                & 0.3001 $\pm$ 0.25         & 0.2897                                      & 169.67  \\  
Ours (OG + DTP) & \underline{0.8680} & \underline{0.2611 $\pm$ 0.24} &  0.2873 &
155.41 \\ \hline 

\end{tabular}}
\end{table}

\begin{figure*}[t!]
\centering
    \begin{subfigure}[t]{0.48\textwidth}
        \centering
        \includegraphics[width=\linewidth]{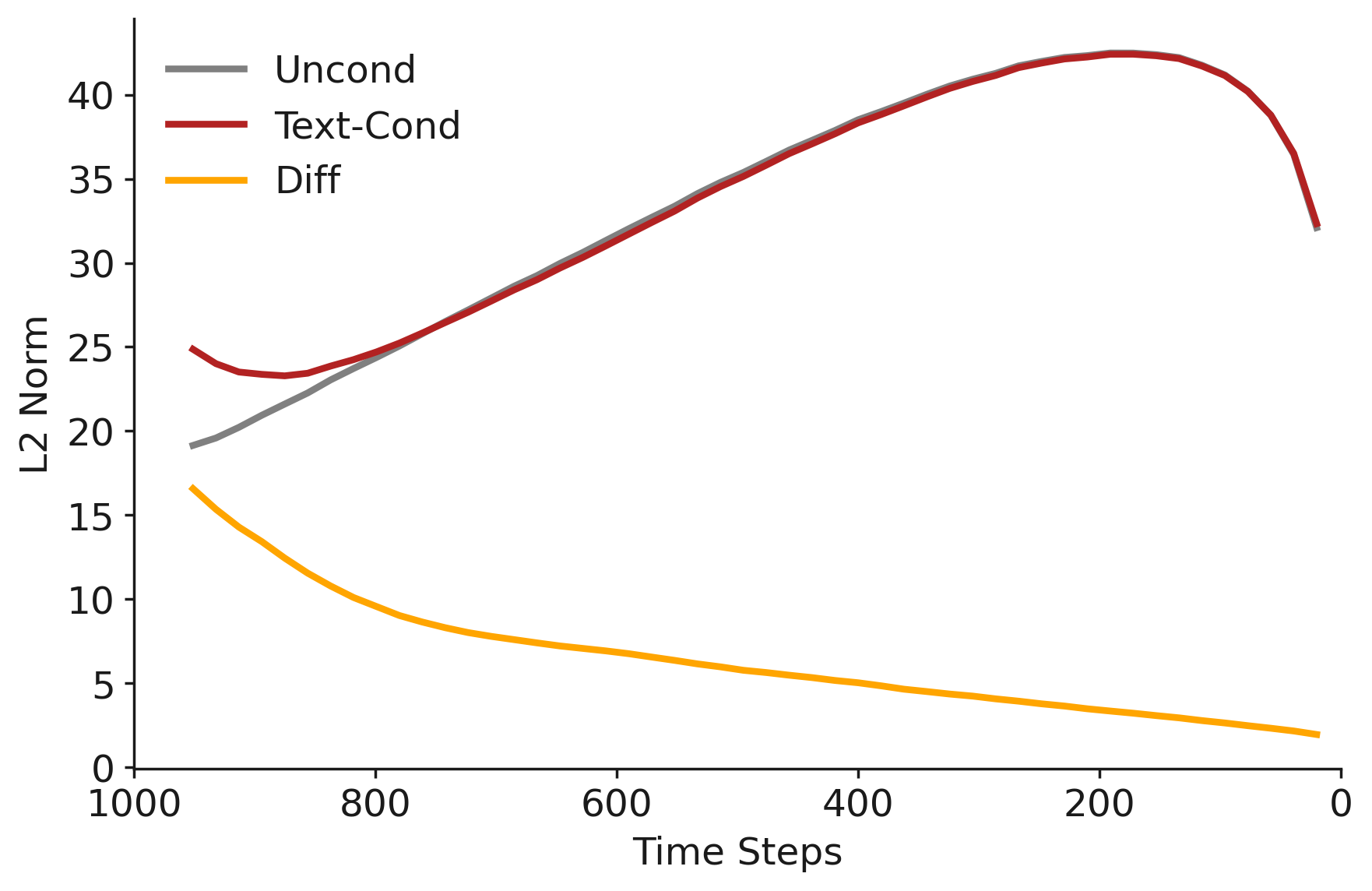}
        \caption{\scriptsize SDv2.1 fine-tuned on LAION-100k }
    \end{subfigure}
    ~
    \begin{subfigure}[t]{0.48\textwidth}
        \centering
        \includegraphics[width=\linewidth]{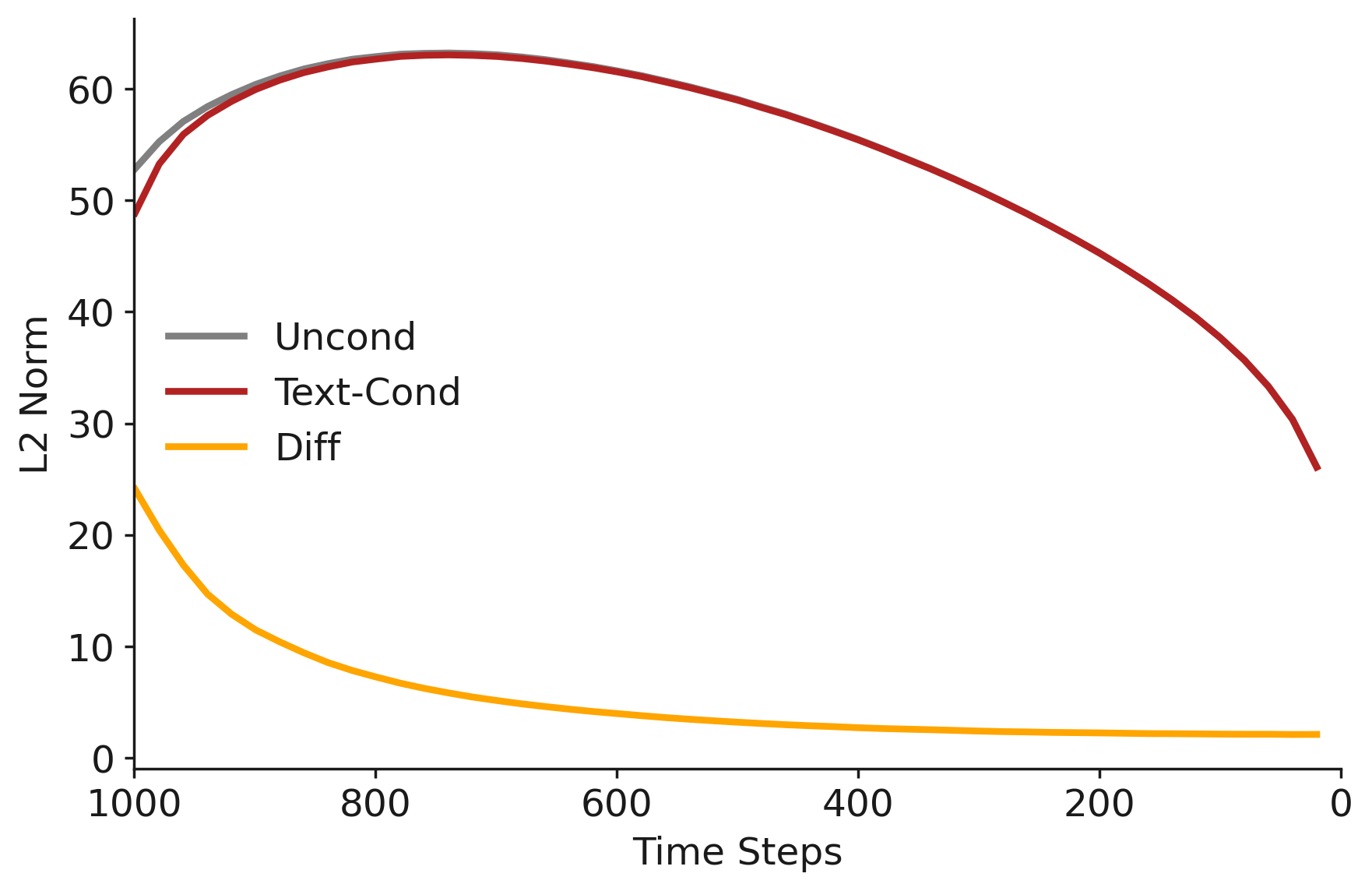}
        \caption{\scriptsize SDv2.1 without fine-tuning }
    \end{subfigure}
    \caption{Plots depicting the trends in the $L_2$ norms of the text-conditioned noise predictions, unconditional noise predictions, and their difference when applying zero CFG during the denoising steps. We see universal transition points appear when SDv2.1 is finetuned on larger datasets as well such as LAION-100k, but this is not visible in the pretrained model. }
    
    \label{fig:l2_norms_sdv1_appendix}
    
\end{figure*}

\section{Results when Finetuning SDv2.1 \\ on LAION-100k}
\label{sec:scenario3_additional_results}

We observed that finetuning SDv2.1 on the LAION-10k dataset \cite{somepalli2023understanding_neurips} leads to exact memorization when using the same text prompts. However, on the LAION-100k dataset, there are stylistic similarities in the output, but verbatim memorization is not always present. In the former scenario, a major proportion of prompts lead to memorized outputs that are extremely similar to the original training images (similarity score $>0.5$). Thus, we choose to focus on mitigating memorization when finetuning on 10,000 samples. However, for both these dataset sizes, the disparity in the text-conditioned and unconditional scores appears in the initial time steps as visualized in Figure \ref{fig:l2_norms_sdv2}(a) for LAION-10k and \ref{fig:l2_norms_sdv1_appendix}(a) for LAION-100k. 

Applying our approach to a SDv2.1 finetuned on the LAION-100k dataset shows similar improvements in similarity scores. We summarize the results in Table \ref{tab:laion-100k}.

\begin{table}
    \centering
        \caption{Results when fine-tuning SDv2.1 on LAION-100k. }
        \label{tab:laion-100k}
        \resizebox{\columnwidth}{!}{
\begin{tabular}{lccc}
\hline 
 & \textbf{Similarity (95pc)} & \textbf{CLIP Score} & \textbf{FID} \\ \hline 
No mitigation                         & 0.3952     &  0.314 & 11.46      \\ \hline
Ours (STP)               & \textbf{0.2861}               & 0.309                         & 16.06              \\ \hline
    \end{tabular}
    }
\end{table}

\section{Detailed Experiment Settings}
\label{appendix:implementation_details}

\subsection{Scenario 1}

We finetune Stable Diffusion v2.1 on 10,000 examples from the LAION dataset, publicly available here\footnote{\href{https://drive.google.com/drive/folders/1TT1x1yT2B-mZNXuQPg7gqAhxN_fWCD__}{https://drive.google.com/drive/folders/1TT1x1yT2B-mZNXuQPg7gqAhxN\_fWCD\_\_}}. The model was finetuned with image sizes 256x256 for 100,000 steps to allow it to memorize the small dataset entirely. During inference, it was observed that the same prompts as the training dataset lead to similar outputs. This memorization scenario was initially studied by Somepalli et al. \cite{somepalli2023understanding_neurips}. We followed the same inference protocol with 50 inference steps using the DPM Multi-step solver \cite{lu2022dpm}.

\subsection{Scenario 2}

We finetuned Stable Diffusion v2.1 on the Imagenette dataset~\cite{Howard_Imagenette_2019} comprising 10 classes of the full ImageNet dataset. These classes are - bench, English springer, cassette player, chain saw, church, French horn, garbage truck, gas pump, golf ball, and parachute. Similar to Scenario 1, this was initially studied by \cite{somepalli2023understanding_neurips}. We finetuned the model for 40,000 steps which led to the best memorization vs image quality trade-off. All images were given the same prompt template, \textit{An image of \{Object\}}, where \textit{Object} is an ImageNet class. During sampling, we used the DPM multi-step solver \cite{lu2022dpm} for 50 inference steps.

\subsection{Scenario 3}

We used the pre-trained model weights from Wen et al. \cite{wen2024detecting} available publicly\footnote{\href{https://drive.google.com/drive/folders/1XiYtYySpTUmS_9OwojNo4rsPbkfCQKBl}{https://drive.google.com/drive/folders/1XiYtYySpTUmS\_9OwojNo4rsPbkfCQKBl}}. The model was finetuned using 200 samples that were duplicated 200 times with an additional 120,000 samples from the LAION dataset. We only expect the 200 duplicated samples to be memorized. During image generation, we used the DPM multi-step solver \cite{lu2022dpm} for 50 inference steps.

\subsection{Scenario 4}

Webster et al. \cite{webster2023reproducible} had found 500 prompts memorized by the pre-trained Stable Diffusion v1.4. We used this prompt dataset directly. Similar to other scenarios for sampling images, we used DPM multi-step solver \cite{lu2022dpm} for 50 inference steps. Since the memorized samples were found in a pre-trained model, the exact cause of memorization is unknown. We observed a large number of template memorization samples as well in addition to verbatim memorization. We observed a strong presence of trigger tokens in these memorized prompts, where the outputs were closely related to either celebrity, movie, or book titles.

\vspace{-0.1in}

\section{Prompts for Figure \ref{fig:qual_comp_laion10k}}
\label{sec:prompts_figure_laion10k}

\begin{itemize}
\item 'Frozen Flower A'
\item  'Clint Eastwood - Camp Pendleton'
\item  'Adult Kids Half Face Rabbit Bunny Mask for Halloween/Easter/Masquerade/Carnival/Party-Luckyfine'
 \item 'Eplans Craftsman House Plan Open Layout With Flex Space'
\item  'How the Mustang got its clothes'
\item  'Spidey and Cap team up against Doctor Doom'
\end{itemize}

\vspace{-0.1in}

\section{Prompts for Figure \ref{fig:qual_comp_200_memorized_images}}
\label{sec:prompts_figure_200_memorized}

\begin{itemize}
    \item Fattoush Salad with Roasted Potatoes
    \item illusion art step by step ; Illusion Kunst, Illusion Art, Illusion Paintings, Coffee Drawing, Coffee Art, Coffee Time, Coffee Shop, Coffee Cups, Pencil Art Drawings
    \item Christmas Comes Early to U.K. Weekly Home Entertainment Chart
    \item James Dean In Black And White Greeting Card by Douglas Simonson
    \item 3D Metal Cornish Harbour Painting
    \item "In this undated photo provided by the New York City Ballet, Robert Fairchild performs in ""In Creases"" by choreographer Justin Peck which is being performed by the New York City Ballet in New York. (AP Photo/New York City Ballet, Paul Kolnik)"
\end{itemize}
\

\begin{figure*}
    \centering
    \begin{subfigure}[t]{0.53\textwidth}
        \includegraphics[width=\linewidth]{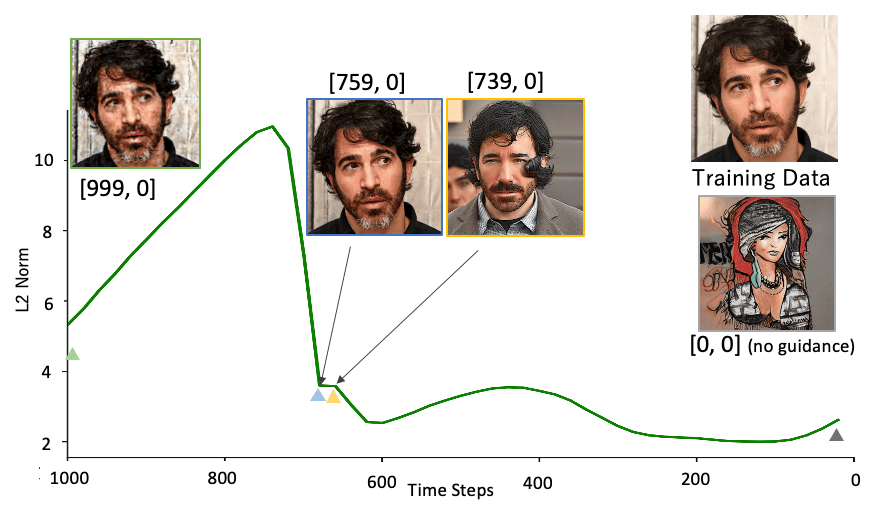}
        \caption{Prompt: Chris Messina In Talks to Star Alongside Ben Affleck in <i>Live By Night</i>}
    \end{subfigure}
    \\
    \begin{subfigure}[t]{0.53\textwidth}
        \includegraphics[width=\linewidth]{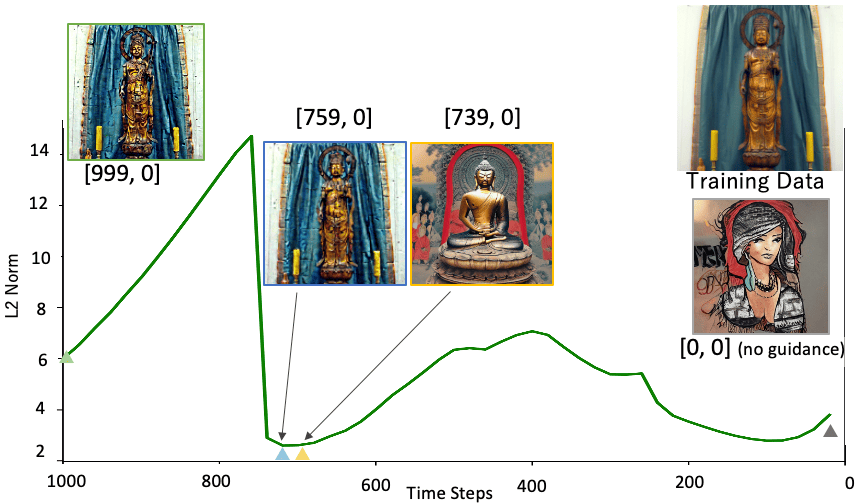}
        \caption{Prompt: Talks on the Precepts and Buddhist Ethics}
    \end{subfigure}
    \\
    \begin{subfigure}[t]{0.53\textwidth}
        \includegraphics[width=\linewidth]{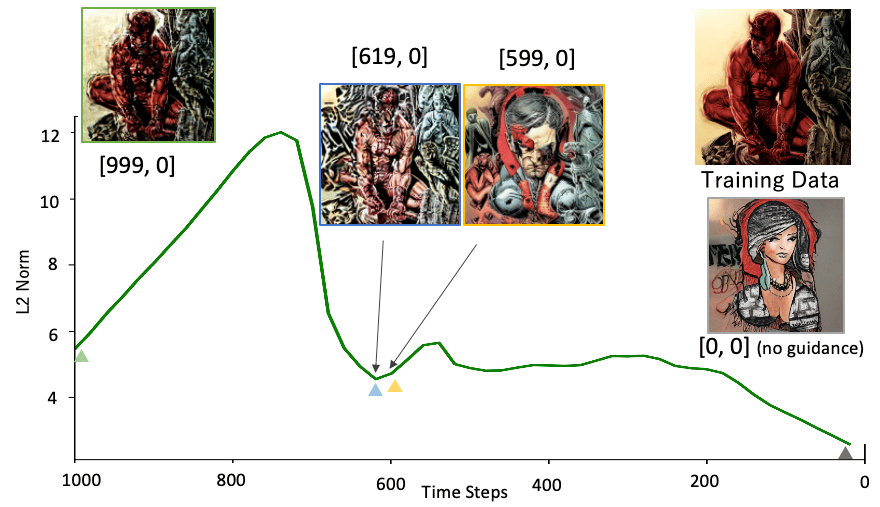}
        \caption{Prompt: As Punisher Joins <i>Daredevil</i> Season Two, Who Will the New Villain Be?}
    \end{subfigure}
    \caption{Examples show the transition of images from memorized to non-memorized if we apply CFG starting from an ideal transition point. }
    \label{fig:chris_melissa}
\end{figure*}

\section{More Examples of Transition Points}

We provide additional visualizations of transition points coinciding with a fall in conditional guidance in Figure \ref{fig:chris_melissa}.

\section{More Visual Examples}
\label{sec:appendix_visual_examples}

\begin{figure*}[t!]
    \centering
    \begin{subfigure}[t]{0.125\textwidth}
        \centering
        \includegraphics[height=6.4in]{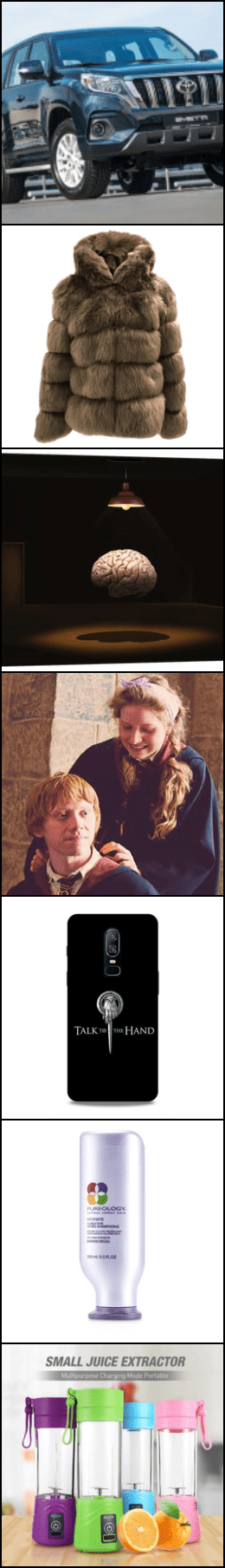}\\
        Original
    \end{subfigure}
    ~ 
    \begin{subfigure}[t]{0.125\textwidth}
        \centering
        \includegraphics[height=6.4in]{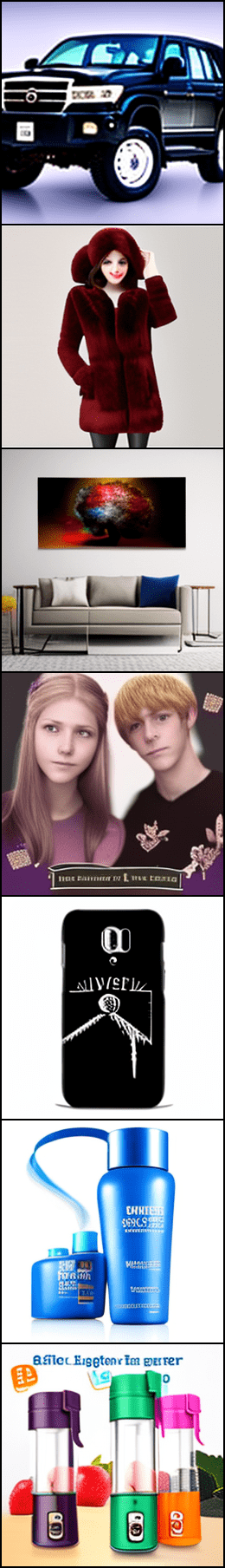}\\
        Add rand numb~\citep{somepalli2023understanding_neurips}
    \end{subfigure}
    ~
    \begin{subfigure}[t]{0.125\textwidth}
        \centering
        \includegraphics[height=6.4in]{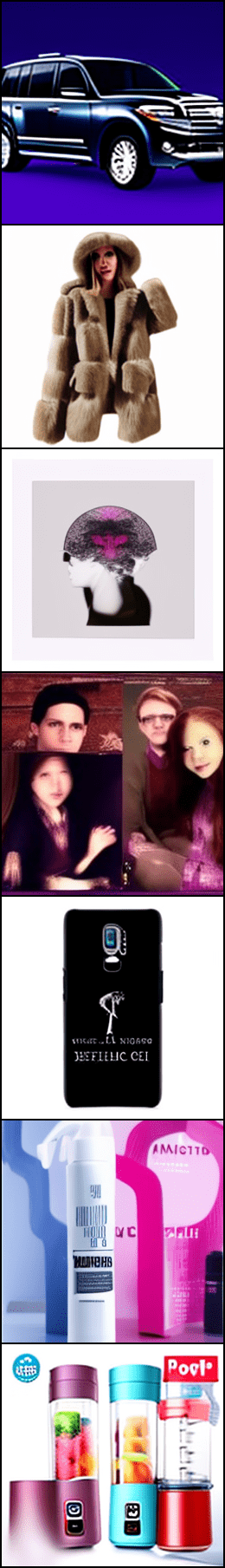}\\
        Add rand word~\citep{somepalli2023understanding_neurips}
    \end{subfigure}
    ~
    \begin{subfigure}[t]{0.125\textwidth}
        \centering
        \includegraphics[height=6.4in]{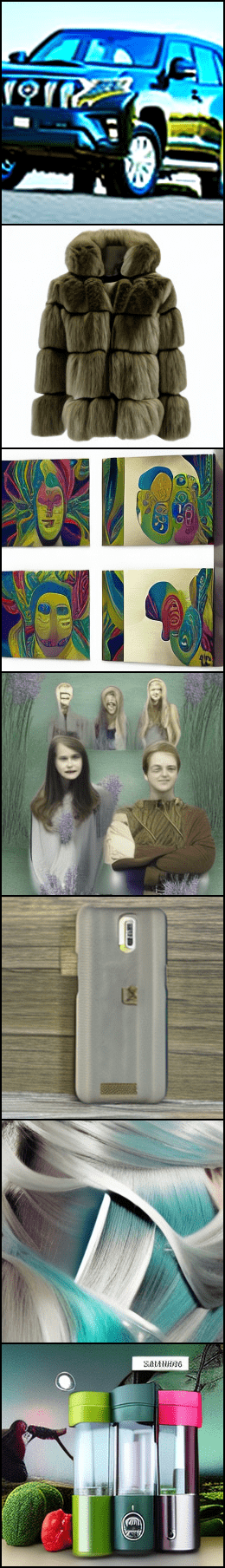}\\
        Wen et al.~\cite{wen2024detecting}
    \end{subfigure}
    ~
    \begin{subfigure}[t]{0.125\textwidth}
        \centering
        \includegraphics[height=6.4in]{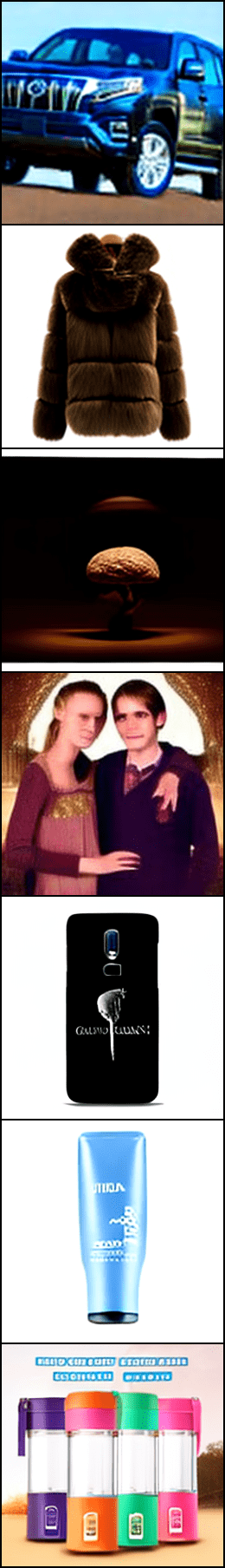}\\
        Ren et al.~\cite{ren2024unveiling}
    \end{subfigure}
    ~
    \begin{subfigure}[t]{0.125\textwidth}
        \centering
        \includegraphics[height=6.4in]{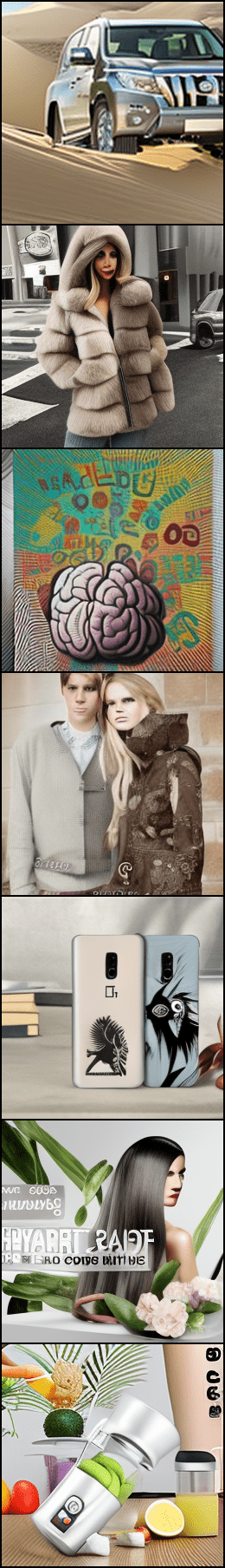}\\
        Ours (STP)
    \end{subfigure}
    ~
    \begin{subfigure}[t]{0.125\textwidth}
        \centering
        \includegraphics[height=6.4in]{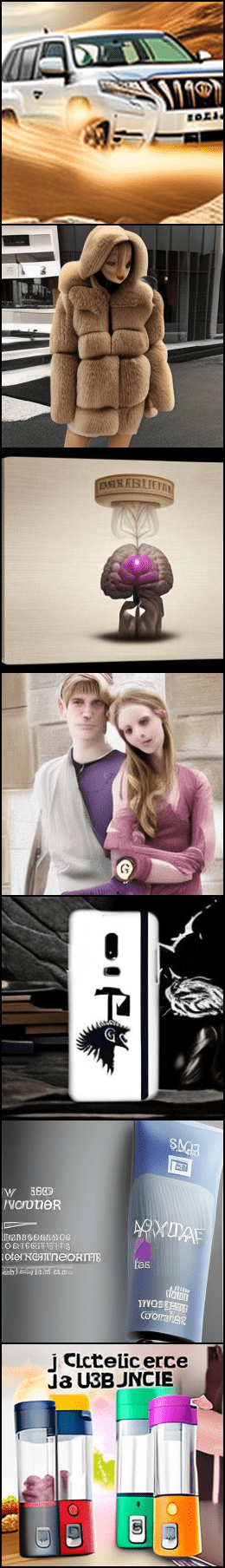}\\
        Ours (OG+STP)
    \end{subfigure}
    \caption{Qualitative results comparing the proposed approach with the baselines in Scenario 1. The following prompts have been used to generate these images: (a) 'Fila Disruptor Animal WMN Zebra/ Black'; (b)
 'Hooded Coat Thicken Fluffy Faux Fur Jacket'; (c)
 'Cerebro (Brain) Canvas Art Print'; (d)
 'couple, lavender brown, and half-blood prince image'; (e)
 'Baby Paintings - Two Mares and a Foal by George Stubbs'; (f)
 'Hydrate Condition (For Dry Colour-Treated Hair)'; (g)
 'Portable USB Electric Juicer'}
    \label{fig:qual_comp_laion10k_appendix}
\end{figure*}

\begin{figure*}
    \centering

    \begin{subfigure}[t]{0.125\textwidth}
        \centering
        \includegraphics[height=6.4in]{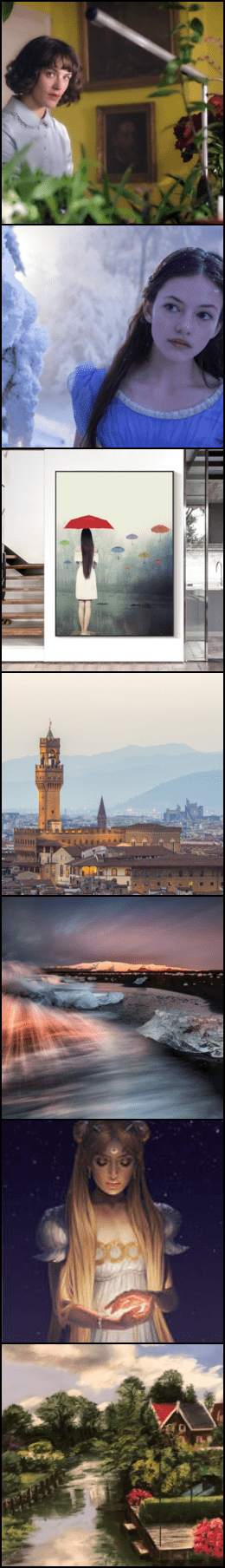}\\
        Training
    \end{subfigure}
    ~ 
    \begin{subfigure}[t]{0.125\textwidth}
        \centering
        \includegraphics[height=6.4in]{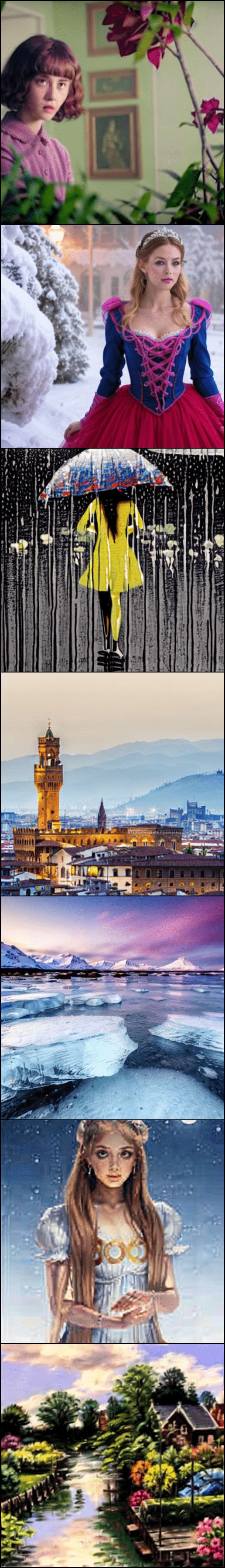}\\
        Add rand numb~\citep{somepalli2023understanding_neurips}
    \end{subfigure}
    ~
    \begin{subfigure}[t]{0.125\textwidth}
        \centering
        \includegraphics[height=6.4in]{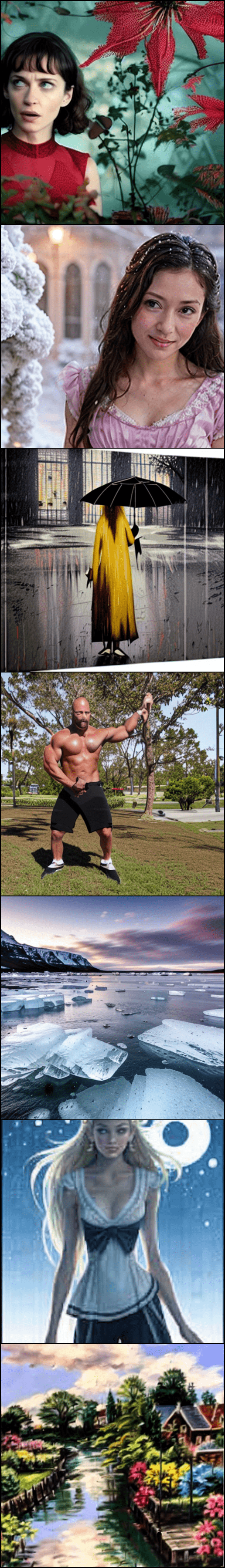}\\
        Add rand word~\citep{somepalli2023understanding_neurips}
    \end{subfigure}
    ~
    \begin{subfigure}[t]{0.125\textwidth}
        \centering
        \includegraphics[height=6.4in]{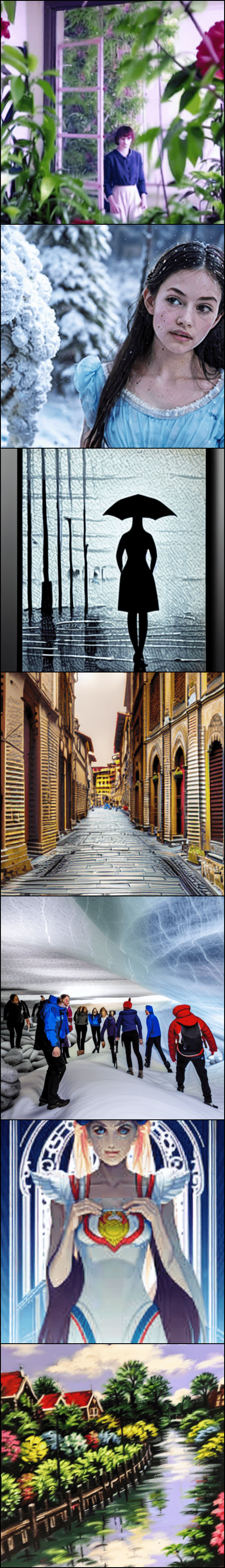}\\
        Wen et al.~\cite{wen2024detecting}
    \end{subfigure}
    ~
    \begin{subfigure}[t]{0.125\textwidth}
        \centering
        \includegraphics[height=6.4in]{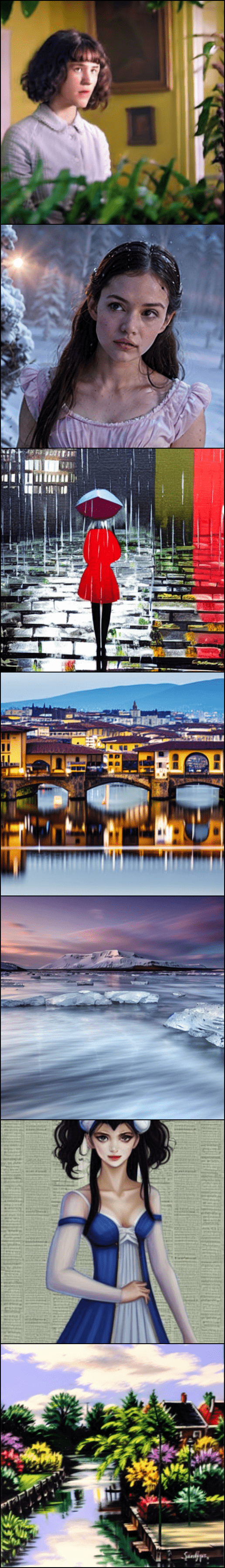}\\
        Ren et al.~\cite{ren2024unveiling}
    \end{subfigure}
    ~
    \begin{subfigure}[t]{0.125\textwidth}
        \centering
        \includegraphics[height=6.4in]{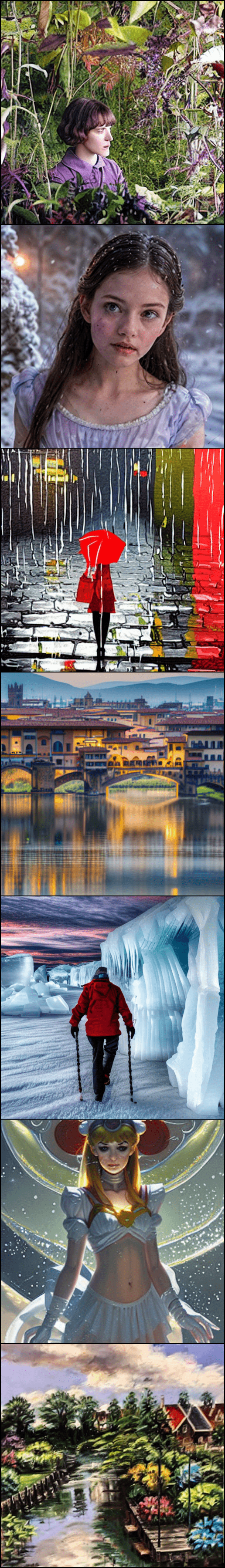}\\
        Ours (DTP)
    \end{subfigure}
    ~
    \begin{subfigure}[t]{0.125\textwidth}
        \centering
        \includegraphics[height=6.4in]{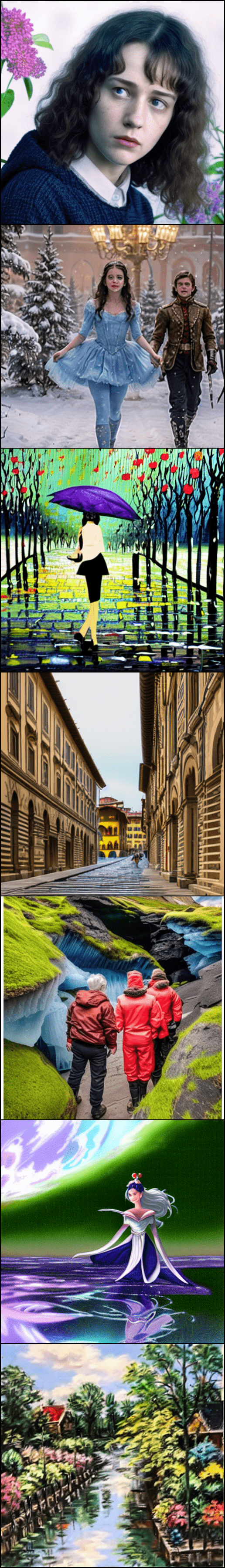}\\
        Ours (OG+DTP)
    \end{subfigure}
    
    \caption{Qualitative results comparing the proposed approach with the baselines in Scenario 3. The following prompts have been used (a) This Beautiful Fantastic (2016); (b)
91079f7c6f6 Disney s The Nutcracker and the Four Realms Movie Review - Theresa s Reviews; (c)
Woman with Umbrella In The Rain Painting Printed on Canvas 1; (d)
View of Florence during the day Stock Photo - 22581191; (e)
Ice Cave Day Tour with Flights from Reykjavik; (f)
Sailor Moon by Charlie-Bowater; (g)
Pastel artwork of a canal in Edam, Netherlands, by Susan Marino. }
    \label{fig:qual_comp_200_memorized_images_appendix}
\end{figure*}

\begin{figure*}[t!]
    \centering
    \begin{subfigure}[t]{0.125\textwidth}
        \centering
        \includegraphics[height=6.4in]{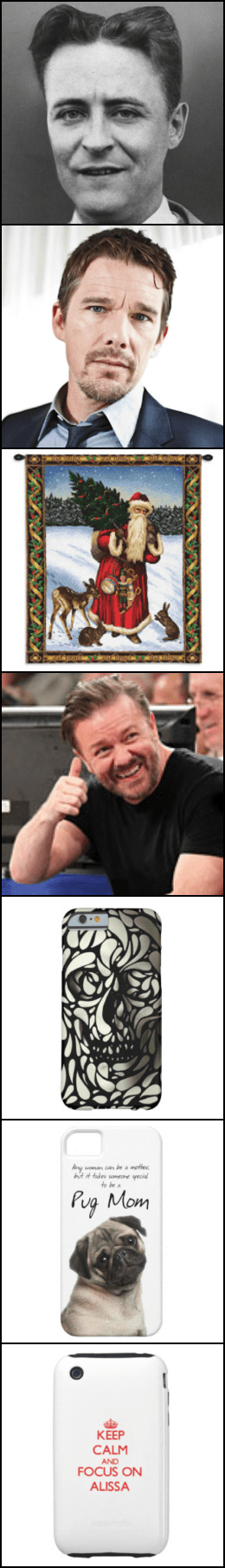}\\
        Training
    \end{subfigure}
    ~ 
    \begin{subfigure}[t]{0.125\textwidth}
        \centering
        \includegraphics[height=6.4in]{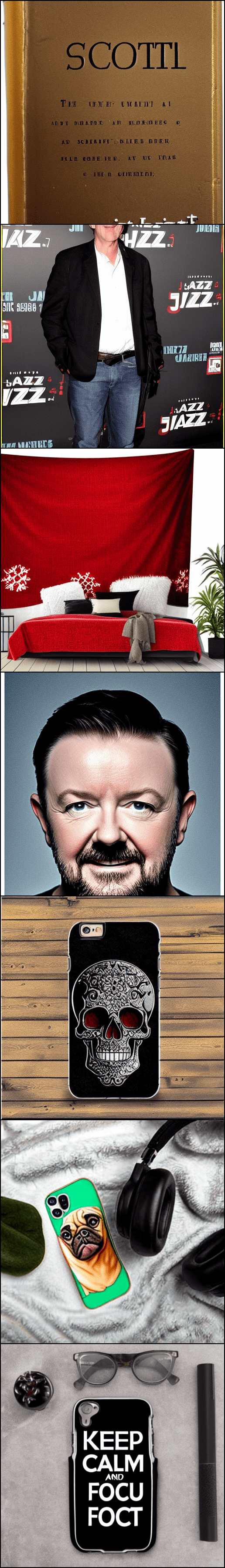}\\
        Add rand numb~\citep{somepalli2023understanding_neurips}
    \end{subfigure}
    ~
    \begin{subfigure}[t]{0.125\textwidth}
        \centering
        \includegraphics[height=6.4in]{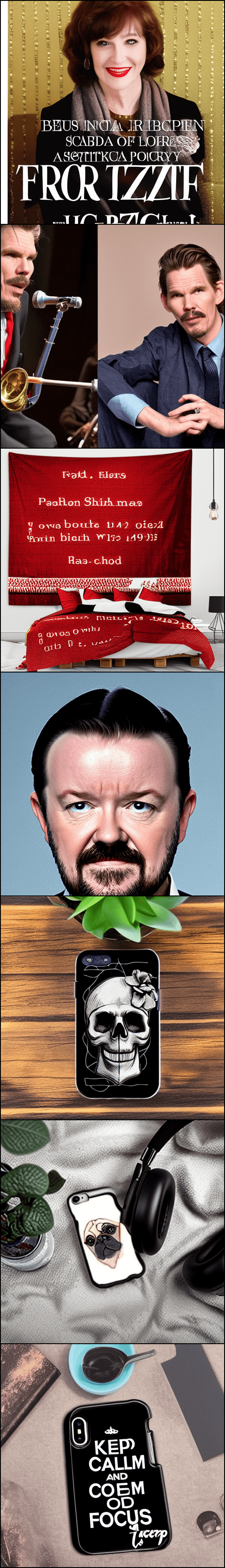}\\
        Add rand word~\citep{somepalli2023understanding_neurips}
    \end{subfigure}
    ~
    \begin{subfigure}[t]{0.125\textwidth}
        \centering
        \includegraphics[height=6.4in]{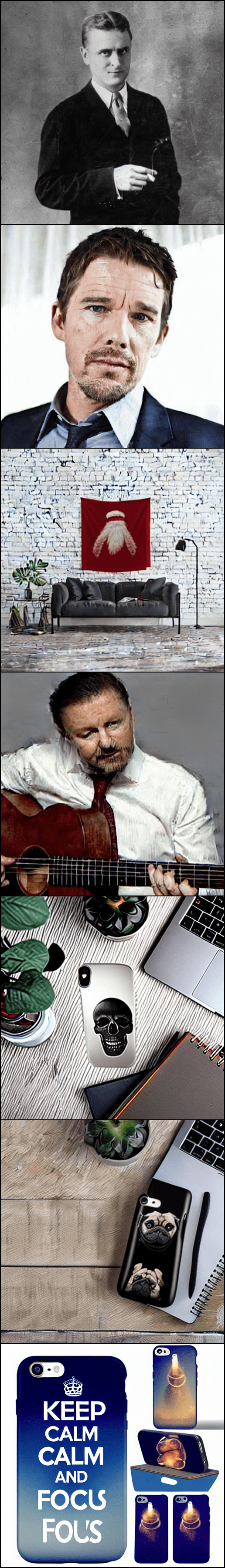}\\
        Wen et al.~\cite{wen2024detecting}
    \end{subfigure}
    ~
    \begin{subfigure}[t]{0.125\textwidth}
        \centering
        \includegraphics[height=6.4in]{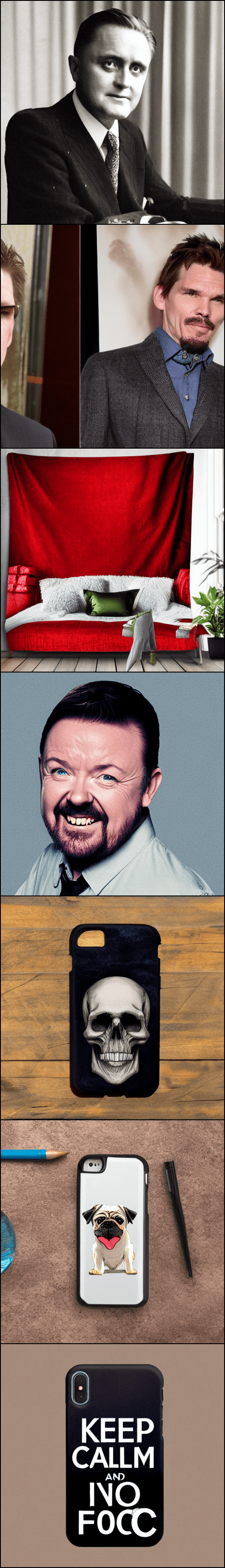}\\
        Ren et al.~\cite{ren2024unveiling}
    \end{subfigure}
    ~
    \begin{subfigure}[t]{0.125\textwidth}
        \centering
        \includegraphics[height=6.4in]{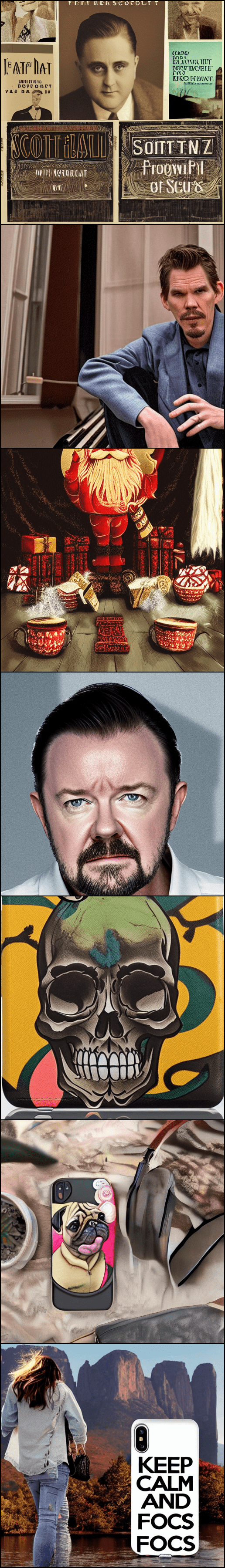}\\
        Ours (DTP)
    \end{subfigure}
    ~
    \begin{subfigure}[t]{0.125\textwidth}
        \centering
        \includegraphics[height=6.4in]{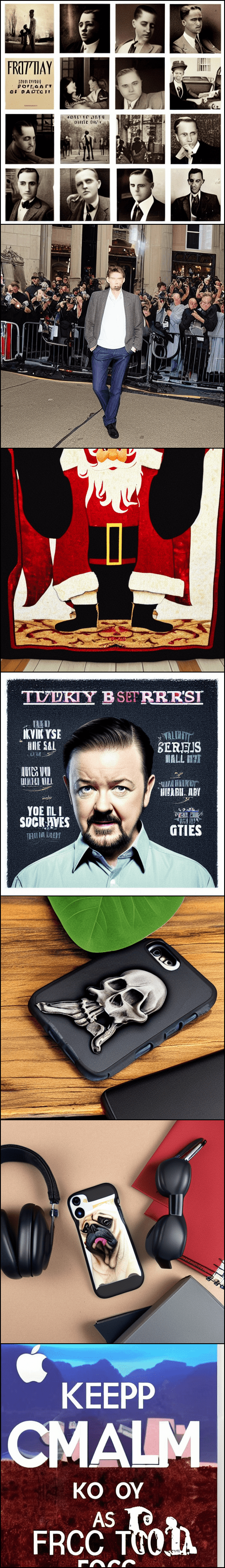}\\
        Ours (OG+DTP)
    \end{subfigure}
    \caption{Qualitative results comparing the proposed approach with the baselines in Scenario 4. The prompts used for generating these images are: (a) Read a Previously Unpublished F. Scott Fitzgerald Story; (b) Ethan Hawke to Star as Jazz Great Chet Baker in New Biopic; (c) Father Christmas Red Wall Tapestry Wall Tapestry; (d) Ricky Gervais Promises More David Brent Gigs; (e) Skull 5 barely there iPhone 6 case; (f) Pug Mom iPhone 5 Case; (g) Keep Calm and focus on Alissa iPhone 3 Tough Cover.}
    \label{fig:qual_comp_500_webster_appendix}
\end{figure*}

\begin{figure*}
    \centering
    \includegraphics[width=\linewidth]{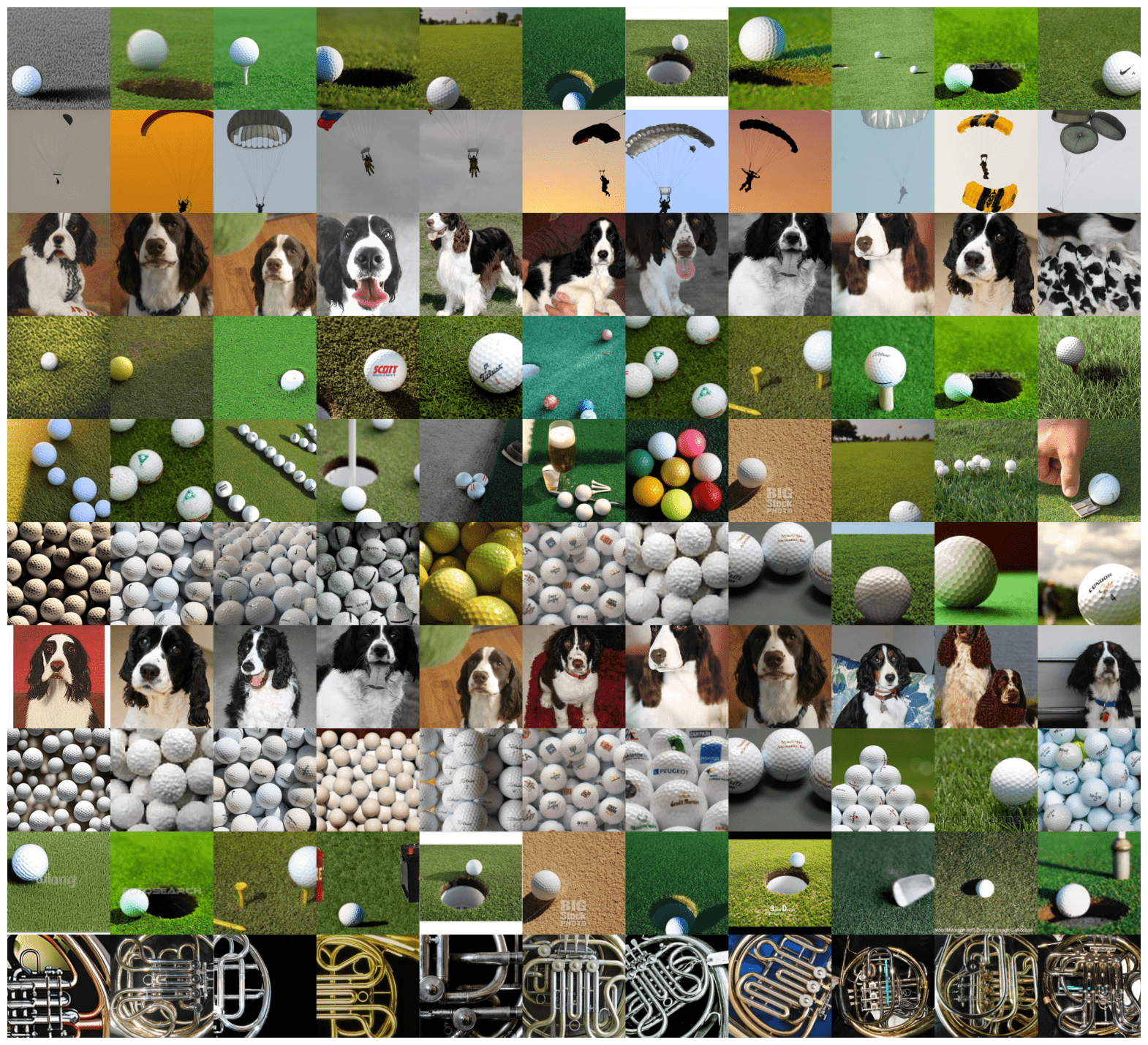}
    \caption{Qualitative results on Scenario 2 showing the closest match of the generated sample (Column 1) with real samples (Columns 2-11) based on the similarity metric. Since exact memorization is not observed for class conditional models, we look at the most similar real images for qualitative results. We did not observe any verbatim memorization.}
    \label{fig:Imagenet}
\end{figure*}

We provide more visual examples comparing our approach with baselines to showcase the effectiveness of our approach in mitigating memorization. We provide examples for Scenario 1 in Figure \ref{fig:qual_comp_laion10k_appendix}, Scenario 2 in Figure \ref{fig:Imagenet}, Scenario 3 in Figure \ref{fig:qual_comp_200_memorized_images_appendix}, and Scenario 4 in Figure \ref{fig:qual_comp_500_webster_appendix}.

\end{document}